%% file: DBULL.tex
\documentclass[review]{elsarticle}

\usepackage{lineno,hyperref}
\usepackage{booktabs}
\usepackage[acronym]{glossaries}
\usepackage{bm}
\usepackage{multirow}
\usepackage{algorithmic,algorithm}
\modulolinenumbers[5]

\journal{Neural Networks}
\input{head_zhu}

\newdefinition{definition}{Definition}

\newacronym{ll}{LL}{Lifelong Learning}
\newacronym{ull}{ULL}{Unsupervised Lifelong Learning}
\newacronym{sll}{SLL}{Supervised Lifelong Learning}
\newacronym{cerr}{CERR}{Cluster Expansion and Redundancy Removal}
\newacronym{dbull}{DBULL}{Deep Bayesian  Unsupervised Lifelong Learning}
\newacronym{vae}{VAE}{Variational Autoencoders}
\newacronym{dpmm}{DPMM}{Dirichlet Process Mixture Model}
\newacronym{curl}{CURL}{Continual Unsupervised Representation Learning}
\newacronym{dnn}{DNNs}{Deep Neural Networks}
\newacronym{dp}{DP}{Dirichlet Process}
\newacronym{mcmc}{MCMC}{Markov Chains Monte Carlo}

\makeatletter
\newcommand{\printfnsymbol}[1]{%
  \textsuperscript{\@fnsymbol{#1}}%
}

\begin{document}
	
	\begin{frontmatter}
		
		\title{Deep Bayesian Unsupervised Lifelong Learning}
	
		\author[department1]{Tingting Zhao*}
		\author[department1]{Zifeng Wang*} 
		\author[department1]{Aria Masoomi}
		\author[department1]{Jennifer Dy}
		
		\address[department1]{Department of Electrical and Computer Engineering, Northeastern University}
		\fntext[myfootnote]{* signifies equal contribution.}
		

		
		
		\begin{abstract}
			\acrfull{ll} refers to the ability to continually learn and solve new problems with incremental available information over time while retaining previous knowledge. Much attention has been given lately to \acrfull{sll} with a stream of labelled data. In contrast, we focus on resolving challenges in \acrfull{ull} with streaming unlabelled data when the data distribution and the unknown class labels evolve over time. Bayesian framework is natural to incorporate past knowledge and sequentially update the belief with new data. We develop a fully Bayesian inference framework for \acrshort{ull} with a novel end-to-end
\acrfull{dbull}
algorithm, which can progressively discover new clusters without forgetting the past with unlabelled data while learning latent representations. To efficiently maintain past knowledge, we develop a novel knowledge preservation mechanism via sufficient statistics of the latent representation for raw data. To detect the potential new clusters on the fly, we develop an automatic cluster discovery and redundancy removal strategy in our inference inspired by Nonparametric Bayesian statistics techniques. We demonstrate the effectiveness of our approach using image and text corpora benchmark datasets in both \acrshort{ll} and batch settings.			
		\end{abstract}
		
		
	\end{frontmatter}
	
	\section{Introduction}
With exposure to a continuous stream of information, human beings are able to learn and discover novel clusters continually by incorporating past knowledge; however, traditional machine learning algorithms mainly focus on static data distributions. Training a model with new information often interferes with previously learned knowledge, which 
typically compromises performance on previous datasets \citep{mccloskey1989catastrophic}.
In order to empower algorithms with the ability of adapting to emerging data while preserving the performance on seen data, a new machine learning paradigm called \acrfull{ll} has recently gained some attention.


\acrshort{ll}, also known as continual learning, was first proposed in \cite{thrun1995lifelong}. It provides a paradigm to exploit past knowledge and learn continually by transferring previously learned knowledge to solve similar but new problems in a dynamic environment, without performance degradation on old tasks. \acrshort{ll} is still an emerging field and most existing research work \citep{thrun1996learning,ruvolo2013ella, chen2015lifelong} has focused on \acrfull{sll}, where the boundaries between different tasks are known and each task refers to a supervised learning problem with output labels provided. The term \emph{task} can have different meanings under various contexts. For example, different tasks can represent different subsets of classes or labels in the same supervised learning problem \citep{sarwar2019incremental} and can also represent supervised learning problems in different fields, where researchers target to perform continual learning across different domains \citep{hou2018lifelong} or lifelong transfer learning \citep{ruvolo2013active,isele2016using}.

While most research on \acrshort{ll} has focused on resolving challenges in \acrshort{sll} problems with class labels provided, we instead 
consider \acrfull{ull} problems, where the learning system is interacting with a non-stationary stream of unlabelled data and the cluster labels  are unknown. One objective of \acrshort{ull} is to discover new clusters by interacting with the environment dynamically and adapting to the changes in the unlabeled data without external supervision or knowledge. To avoid confusion, it is worth pointing out that our work assumes that the non-stationary streaming unlabelled data come from a single domain;
and we target to develop a single dynamic model that can perform well on all sequential data at the end of each training stage without forgetting previous knowledge. This setting can serve as a reasonable starting point for \acrshort{ull}. We leave the challenges in \acrshort{ull} across different problem domains for future work.


To retain knowledge from past data and learn new clusters from new data continually, good representations of the raw data make it easier to extract information and, in turn, support effective learning. Thus, it is more computationally appealing if we can discover new clusters in a low-dimensional latent space instead of the complex original data space while performing representation learning. To achieve this, we propose  \acrfull{dbull},
which is a flexible probabilistic generative model that can adapt to new data and expand with new clusters while seamlessly learning deep representations in a Bayesian framework. 

 A critical objective of \acrshort{ll} is to achieve consistently good performance incrementally as new data arrive in a streaming fashion, without performance decrease on previous data, even if the data may have been completely overwritten. Compared with a traditional batch learning setting, there are additional {challenges} to resolve in a \acrshort{ll} setting. One important question is {how to design a knowledge preservation scheme} to efficiently maintain previously learned information. To discover the new clusters automatically with streaming, unlabelled data, another challenge is { how to design a dynamic model that can expand} with incoming data to perform unsupervised learning. The last challenge is {how to design an end-to-end inference algorithm} to obtain good performance {in an incremental learning way}. To answer these questions, we make the following contributions:
\begin{itemize}
\setlength{\parskip}{0pt}
\setlength{\itemsep}{0pt plus 1pt}
\item To solve the  challenges in \acrshort{ull}, we provide a fully Bayesian formulation that performs representation learning, clustering and automatic new cluster discovery simultaneously via our end-to-end novel variational inference strategy \acrshort{dbull}.
\item To efficiently extract and maintain knowledge seen in earlier data, we provide innovation in our incremental inference strategy by first using sufficient statistics in the latent space in an \acrshort{ll} context. 
\item To discover new clusters in the emerging data, we choose a nonparametric Bayesian prior to allow the model to grow dynamically.
We develop a sequential Bayesian inference strategy to perform representation learning simultaneously with our proposed \acrfull{cerr} trick to discover new clusters on the fly without imposing bounds on the number of clusters, unlike most existing algorithms, using a truncated \acrfull{dp} \citep{blei2006variational}.  

\item To show the effectiveness of \acrshort{dbull}, we conduct experiments on image and text benchmarks. \acrshort{dbull} can achieve superior performance compared with state-of-the-art methods in both
\acrshort{ull}  and 
classical batch settings.
\end{itemize}
	
\section{Related Work} 
\subsection{Alleviating Catastrophic Forgetting in Lifelong Learning}

Research on \acrshort{ll} aims
to learn knowledge in a continual fashion without performance degradation on previous tasks when trained for new data. Reference  \cite{parisi2019continual} have provided a comprehensive review on \acrshort{ll} with neural networks. The main challenge of \acrshort{ll} using \acrfull{dnn} is that they often suffer from a phenomenon called \emph{catastrophic forgetting} or \emph{catastrophic interference}, where 
a model's performance on previous tasks may decrease abruptly due to the interference of training with new information \citep{mccloskey1989catastrophic, mcclelland1995there}. Recent work aims at adapting a learned model to new information while ensuring the performance on previous data does not decrease. Currently, there are no universal past knowledge preservation schemes for different algorithms and settings \citep{chen2018lifelong} in \acrshort{ll}. Regularization methods target to reduce interference of new learning by minimizing changes to certain parameters that are important to previous learning tasks \citep{kirkpatrick2017overcoming, zenke2017continual}. Alternative approaches based on rehearsal have also been proposed to alleviate catastrophic forgetting while training \acrshort{dnn} sequentially. Rehearsal methods use either past data \citep{robins1995catastrophic, xu2018lifelong}, coreset data summarization \citep{nguyen2017variational} or a generative model \citep{shin2017continual} to capture the data distribution of previously seen data. 

However, most of the existing \acrshort{ll} methods focus on supervised learning tasks \citep{kirkpatrick2017overcoming,nguyen2017variational, hou2018lifelong, shin2017continual}, where each learning task performs supervised learning with output labels provided. In comparison, we propose a novel  alternative knowledge preservation scheme in an unsupervised learning context via { sufficient statistics}.  This is in contrast to existing work which uses previous model parameters \citep{chen2015lifelong, shu2016lifelong}, representative items exacted from previous models \citep{shu2017lifelong} or past raw data, or coreset data summarization \citep{robins1995catastrophic, xu2018lifelong, nguyen2017variational} as previous knowledge for new tasks. Our proposal to use sufficient statistics is novel and has the advantage of preserving past knowledge without the need of storing previous data while allowing incremental updates as new data arrive by taking advantage of the additive property of sufficient statistics.

\subsection{Comparable Methods in Unsupervised Lifelong Learning}
Recently,  \cite{rao2019continual} proposed \acrfull{curl} to deal with a fully \acrshort{ull} setting with unknown cluster labels. We have developed our idea independently in parallel with \acrshort{curl} but in a fully Bayesian framework. \acrshort{curl} is the most related and comparable method to ours in the  literature. \acrshort{curl} focuses on learning representations and discovering new clusters using a threshold method. One major drawback of \acrshort{curl} is that it has over-clustering issues as shown in their real data experiment. We also show this empirically and demonstrate the improvement of our method over \acrshort{curl} in our experiment section. In contrast to \acrshort{curl}, we provide a novel probabilistic framework with a nonparametric Bayesian prior to allow the model to expand without bound automatically instead of using an ad hoc threshold method as in \acrshort{curl}. We develop a novel end-to-end variational inference strategy for learning deep representations and detecting novel clusters in \acrshort{ull} simultaneously. 

\subsection{Bayesian Lifelong Learning}
A Bayesian formulation is a natural choice for \acrshort{ll} since it provides a systematic way to incorporate previously learnt information in the prior distribution and obtain a posterior distribution that combines both prior belief and new information. The  sequential  nature  of Bayes theorem also paves the way to recursively update an approximation of the posterior distribution and then use it as a new prior to guide the learning for new data in \acrshort{ll}. 

In \cite{nguyen2017variational}, the authors propose a Bayesian formulation in \acrshort{ll}. Although both under a Bayesian framework, our work is different from \cite{nguyen2017variational} due to different objectives, inference strategies and knowledge preservation techniques. In \cite{nguyen2017variational}, the authors provide a variational online inference framework for deep discriminative models and deep generative models, where they studied the approximate posterior distribution of the parameters in \acrshort{dnn} in a continual fashion. However, their method does not have the capacity to find out the latent clustering structure of the data or detect new clusters for emerging data. In contrast, we develop a novel Bayesian framework for representation learning and discovering latent clustering structure and new clusters on the fly together with a novel end-to-end variational inference strategy in an \acrshort{ull} context. 

\subsection{Deep Generative Unsupervised Learning Methods in a Batch Setting}
Recent research  has focused on combining deep generative models to learn good representations of the original data and conduct clustering analysis in an unsupervised learning context \citep{kingma2014auto, johnson2016composing,xie2016unsupervised,jiang2017variational,goyal2017nonparametric}. However, the latest existing methods are designed for an independent and identically distributed (i.i.d.) batch training mode instead of a \acrshort{ll} context. The majority of these methods are in a static unsupervised learning setting, where the number of clusters are fixed in advance. Thus, these methods cannot detect potential new clusters when new data arrive or the data distribution changes. These methods cannot adapt to a \acrshort{ll} setting. 

To summarize, our work fills the gap by providing a fully Bayesian framework for \acrshort{ull}, which has the unique capacity to use a deep generative model for representation learning while performing new cluster discovery on the fly with a nonparametric Bayesian prior and our proposed \acrshort{cerr} technique. To alleviate catastrophic forgetting challenge in \acrshort{ll}, we propose to use sufficient statistics to maintain knowledge as a novel alternative to existing methods. We further develop an  end-to-end Bayesian inference strategy \acrshort{dbull} to achieve our goal. 
	
\section{Model} 
\subsection{Problem Formulation}
In our \acrshort{ull} setting, a sequence of datasets $D_1$, $D_2, \ldots, D_N$ arrive in a streaming order. When a new dataset $D_N$ arrives in memory, the previous dataset $D_{N-1}$ is no longer available. Our goal is to automatically learn the clusters (unlabeled classes) in each dataset. 

Let $\bm{x}\in X$ represent the unlabeled observation of the current dataset in memory, where $X$ can be a high-dimensional data space. We assume that a low-dimensional latent representation $\bm{z}$ can be learned from $\bm{x}$ and in turn can be used to reconstruct $\bm{x}$. We assume that the variation among observations $\bm{x}$ can be captured by its latent representation $\bm{z}$. Thus, we let $\bm{y}$ represent the unknown cluster membership of $\bm{z}$ for observation $\bm{x}$.

We target to find: (1) a good low-dimensional latent representation $\bm{z}$ from $\bm{x}$ to efficiently extract knowledge from the original data; (2) the clustering structure within the new dataset with the capacity to discover potentially novel clusters without forgetting the previously learned clusters of the seen datasets; and, (3) an incremental learning strategy to optimize the cluster learning performance for a new dataset without dramatically degrading the clustering  performance in seen datasets.

\begin{figure}[htbp]
\centerline{\includegraphics[scale=0.45]{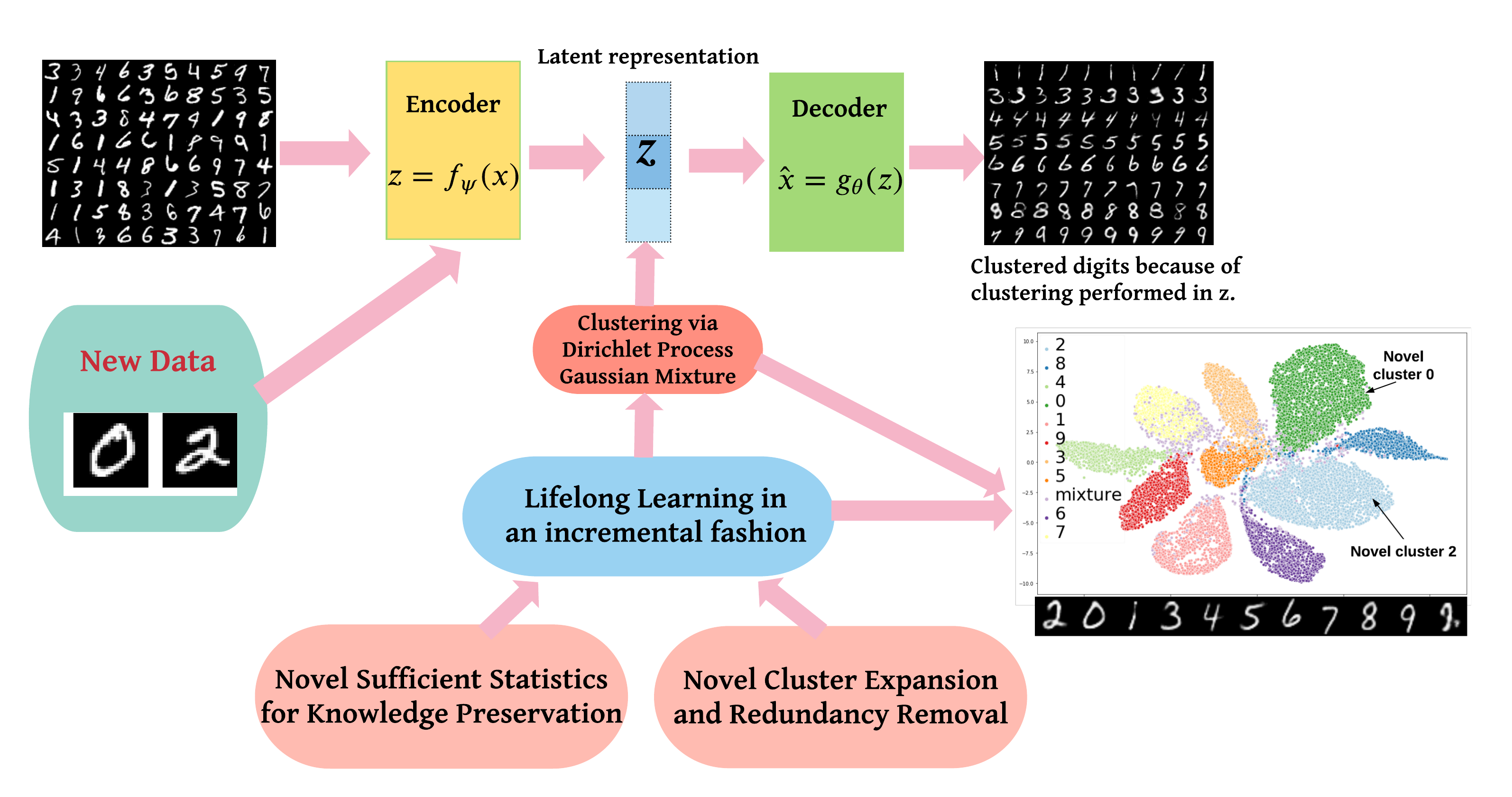}}
\caption{Flow chart of DBULL, best viewed in color. Given observations $\bm{x}$,  \acrshort{dbull} learns its latent representation $\bm{z}$ via an encoder $\bm{z}=f_\psi(\bm{x})$ and performs clustering under a Dirichlet Process Gaussian mixture model and reconstruct the original observation via a decoder $\hat{x}=g_\theta(\bm{z})$, where $\psi$ and $\theta$ denote the parameters in the encoder and decoder respectively. To perform \emph{Lifelong Learning} in an incremental fashion when dealing with streaming data, we introduce two novel components: \emph{Sufficient Statistics} for knowledge preservation and \emph{Cluster Expansion and Redundancy Removal} to create and merge clusters.}
\label{fig:flow}
\end{figure}

\begin{figure}[htbp]
\centerline{\includegraphics[scale=0.55]{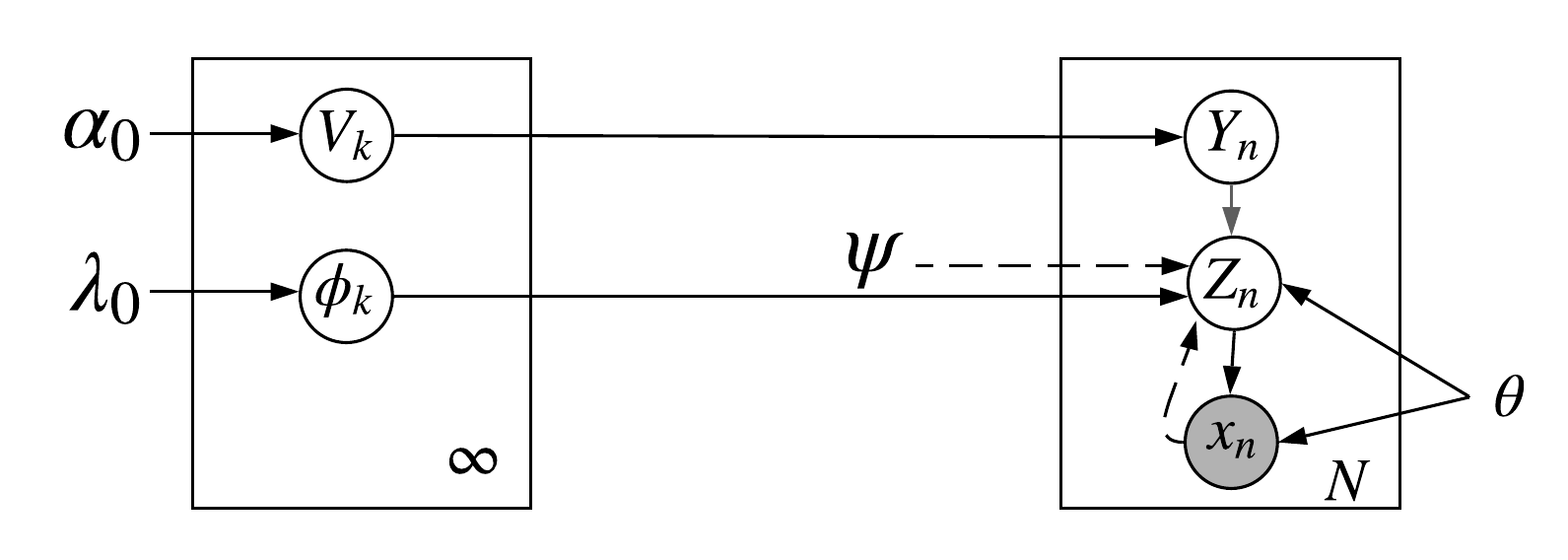}}
\caption{Graphical model representation of DBULL. Nodes denote random variables, edges denote possible dependence, and plates denote replication. Solid lines denote the generative model; Dashed lines denote the variational approximation.}
\label{fig:graphical}
\end{figure}

We summarize our work in a flow chart in Fig.~\ref{fig:flow}, provide its graphical model representation in Fig~\ref{fig:graphical}, and describe the generative process of our graphical model for \acrshort{dbull} in the next section.

\subsection{Generative Process of DBULL}\label{sec:gp}

The generative process for \acrshort{dbull} is as follows. 
 \begin{itemize}
\setlength{\parskip}{0pt}
\setlength{\itemsep}{0pt plus 1pt}
        \item[(a)] Draw a latent cluster  membership $\bm{y} \sim \mbox{Cat}(\pi(\bm{v}))$, where the vector $\bm{v}$ comes from the stick-breaking construction of a Dirichlet Process (DP).
    \item[(b)] Draw a latent representation vector $\bm{z}|\bm{y}=k\sim \mathcal{N}\left(\bm{\mu}_k^*, \bm{\sigma}^{*2}_k\bm{I}\right)$, where $k$ is the cluster membership sampled from (a). 
   \item[(c)] Generate data $\bm{x}$ from $\bm{x}|\bm{z}=\bm{z} \sim \mathcal{N}\left(\bm{\mu}(\bm{z}; \theta), \textrm{diag}(\bm{\sigma}^2(\bm{z}; \theta))\right)$ in the original data space.
   \end{itemize}
   
In (a), $\textrm{Cat}(\pi(\bm{v}))$ is the categorical distribution parameterized by $\pi(\bm{v})$, where we denote the $k$th element of $\pi(\bm{v})$ as $\pi_k(\bm{v})$, which is the probability for cluster $k$. The value of $\pi(\bm{v})$ depends on a vector of scalars $\bm{v}$ coming from the stick-breaking construction of a \acrfull{dp} \citep{sethuraman1982convergence} and we describe an iterative process to draw $\pi_k(\bm{v})$ in Section~\ref{sec:dp}. \acrshort{curl} uses a latent mixture of Gaussian components to capture the clustering structure in an unsupervised learning context. In comparison, we adopt the DP mixture model in the latent space with the advantages that the number of mixture components can be random and grow without bound as new data arrive, which is an appealing property desired by \acrshort{ll}. 
We further explain in Section~\ref{sec:dp} why DP is an appropriate prior for our problem in details. In (b), $\bm{z}$ is considered a low-dimensional latent representation of the original data $\bm{x}$. We describe in Section~\ref{sec:dp} that a DP Gaussian mixture is used for modelling $\bm{z}$ since it is often assumed that the variation in $\bm{z}$ is able to reflect the variation within $\bm{x}$. The current representation in (b) is for easy understanding. In (c), we assume that the generative model $p_\theta(\bm{x}|\bm{z})$ is parameterized by $g_\theta: Z\rightarrow X$ and $g_\theta(\bm{z})=\left(\bm{\mu}(\bm{z}; \theta), \bm{\sigma}^2(\bm{z}; \theta)\right)$, where $g_\theta$ is chosen as \acrshort{dnn} due to its powerful function approximation and good feature learning capabilities \citep{hornik1991approximation, kingma2014semi, nalisnick2016approximate}.
   
Under this generative process, the joint probability density function can be factorized as 
\begin{align}
\label{eq:joint}
&p(\bm{x}, \bm{y}, \bm{z}, \bm{\phi}, \bm{v})=p_\theta(\bm{x}|\bm{z})p(\bm{z}\vert \bm{y})p(\bm{y}|\pi(\bm{v}))p(\bm{v})p(\bm{\phi}),
\end{align}
where $\bm{\phi}=(\bm{\phi}_1, \bm{\phi}_2, \ldots, \bm{\phi}_k)$ and  $\bm{\phi}_k=(\bm{\mu}_k^*, \bm{\sigma}^{*2}_k)$ represents the parameters of the $k$th mixture component (or cluster), $p(\bm{v})$ and $p(\bm{\phi})$ represent the prior distribution for $\bm{v}$ and $\bm{\phi}$.

Next, we discuss how to choose appropriate $p(\bm{v})$ and $p(\bm{\phi})$ to endow our model with the flexibility to grow the number of mixture components without bound with new data in a \acrshort{ll} setting.

\section{Why Bayesian for DBULL}
In this section, we illustrate why Bayesian framework is a natural choice for our \acrshort{ull} setting.  Recall that we have a sequence of datasets $D_1, D_2, \ldots, D_N$ from a single domain arriving in a streaming order. To mimic a \acrshort{ll} setting, we assume each time only one dataset can fit in memory. One key question in \acrshort{ll} is how to efficiently maintain past knowledge to guide future learning.

\subsection{Bayesian Reasoning for Lifelong Learning}
Bayesian framework is a suitable solution to this type of learning since it learns a posterior distribution, or an approximation of a posterior distribution, that takes advantage of both the prior belief and the additional information in the new dataset. The sequential nature of Bayes theorem ensures valid recursive updates on an approximation of the posterior distribution given the observations. Later, the approximation of the posterior distribution serves as the new prior to guide future learning for new data in \acrshort{ll}. Before describing our inference strategy, we first explain why utilizing the Bayesian updating rule is valid for our problem.

Given $(N-1)$ datasets $\mathcal{D}_{i}$, where $i=1, 2, \ldots, (N-1)$, the posterior after considering the $N$th dataset is
\begin{align}&P(\bm{y}, \bm{z}, \bm{\phi}, \bm{v}|\mathcal{D}_1, \mathcal{D}_2, \ldots, \mathcal{D}_{N})\notag\\
&\propto P(\mathcal{D}_N|\bm{y}, \bm{z}, \bm{\phi}, \bm{v})P(\bm{y}, \bm{z}, \bm{\phi}, \bm{v}|\mathcal{D}_1,\ldots, \mathcal{D}_{N-1}),
 \label{eq:bayesian}
\end{align}
which reflects that the posterior of $(N-1)$ tasks and datasets can be considered as the prior for the next task and dataset. If we know exactly the normalizing constant for $P(\bm{y}, \bm{z}, \bm{\phi}, \bm{v}|\mathcal{D}_1, \mathcal{D}_2, \ldots, \mathcal{D}_{N})$ and
$P(\bm{y}, \bm{z}, \bm{\phi}, \bm{v}|\mathcal{D}_1, \mathcal{D}_2, \ldots, \mathcal{D}_{N-1})$, repeatedly updating \eqref{eq:bayesian} is streaming without the need of reusing past data. However, it is often intractable to compute the normalizing constant exactly. Thus, an approximation of the posterior distribution is necessary to update \eqref{eq:bayesian} since the exact posterior is infeasible to obtain. 

\subsection{Dirichlet Process  Prior}\label{sec:dp}
The \acrshort{dp} is  often used as a nonparametric prior for partitioning exchangeable observations into discrete clusters. The \acrshort{dp} mixture is a flexible mixture model where the number of mixture components can be random and grow without bound as more data arrive. These properties make it a natural  choice for our \acrshort{ll} setting. In practice, we show in our inference   how we expand and merge the number of mixture components as new data arrive by starting from only one cluster in Section~\ref{sec:expansion}. Next, we briefly review \acrshort{dp} and introduce our \acrshort{dp} Gaussian mixture model to derive the joint probability density $p(\bm{x}, \bm{y}, \bm{z}, \bm{\phi}, \bm{v})$ defined in \eqref{eq:joint}. 

A \acrshort{dp} is characterized by a base distribution $G_0$ and a parameter $\alpha$ denoted as $\textrm{DP}(G_0, \alpha)$. A constructive definition of \acrshort{dp} via a stick-breaking process is of the form $G(\cdot)=\sum_{k=1}^{\infty}\pi_k \delta_{\bm{\phi}_k}$,
where $\delta_{\bm{\phi}_k}$ is a discrete measure concentrated at $\bm{\phi}_k\sim G_0$, which is a random sample from the base distribution $G_0$  with mixing proportion $\pi_k$ \citep{ishwaran2001gibbs}. In \acrshort{dp}, the $\pi_k$s are random weights independent of $G_0$ but satisfy $0\leqslant \pi_k\leqslant 1$ and $\sum_{k=1}^{\infty} \pi_k=1$. The weights $\pi_k$ can be drawn through an iterative process:
\[
\pi_k=\begin{cases}
v_1, & \textrm{if}\quad k=1, \\
v_k\prod_{j=1}^{k-1}(1-v_j), & \textrm{for}\quad k>1,
\end{cases}
\]
where $v_k\sim \textrm{Beta}(1, \alpha)$.

Under the generative process of \acrshort{dbull} in Section~\ref{sec:gp}, these $\pi_k$s represent the probabilities for each cluster (mixture component) used in step (a) and $\bm{\phi}_k$ can be seen as the parameters of the Gaussian mixture for $\bm{z}$ in step (b). Thus, given our generative process, the corresponding joint probability density for our model is 
\begin{align}
    p(\bm{x}, \bm{y}, \bm{z}, \bm{\phi}, \bm{v})&=p(\bm{x}|\bm{z};\theta)p(\bm{z}\vert \bm{y})p(\bm{y}|\bm{v})p(\bm{v})G_0(\bm{\phi}|\bm{\lambda}_0)\notag\\
    &=\prod_{n=1}^N \mathcal{N}(\bm{x}_n|\bm{\mu}(\bm{z}_n;\theta),\mathrm{diag}(\bm{\sigma}^2(\bm{z}_n, \theta)))\notag\\
    &\quad\,\prod_{k=1}^{\infty}\mathcal{N}(\bm{z}_n|\bm{\mu}_{k}^*, \bm{\sigma}_k^{*2}\bm{I})P(\bm{y}_n=k|\pi(\bm{v}))\notag\\
    &\quad\quad\quad \mbox{Beta}(v_k|1, \alpha_0)G_0(\bm{\phi}_{k}|\lambda_0).
    \label{eq:pdf}
\end{align}
For a Gaussian mixture model, the base distribution is often chosen as the Normal-Wishart (NW) denoted as $G_0=\mbox{NW}(\bm{\lambda}_0)$ to generate the mixture  parameters $\bm{\phi}_k=\left(\bm{\mu}_k^*, \bm{\sigma}_k^*\right)\sim\mbox{NW}(\bm{\lambda}_0)$, where $\bm{\lambda}_0=(m_0, \beta_0, \nu_0, W_0)=(\bm{0}, 0.2, D+2, I_{D\times D})$, and $D$ is the dimension of the latent vector $\bm{z}$. The values of the hyper-parameter $\bm{\lambda}_0$ are conventional choices in the Bayesian nonparameteric literature for Gaussian mixture. Moreover, the performance of our method is robust to the hyper-parameter values.
	
\section{Inference for DBULL}\label{sec:inference}
There are several new challenges to develop an end-to-end inference algorithm for our problem under the \acrshort{ull}  setting  compared with the batch setting: one has to deal with catastrophic forgetting,  mechanisms for past knowledge preservation, and dynamic model expansion capacity for novel cluster discovery. 
For pedagogical reasons, we first describe our general parameter learning strategy via variational inference for DBULL in a standard batch setting. 
We then describe how we resolve the additional challenges in the lifelong (streaming) learning setting. We describe our novel components in the inference algorithm in terms of a new knowledge preservation scheme via sufficient statistics in Sections~\ref{sec:suff} and an automatic \acrshort{cerr} strategy in  Section~\ref{sec:expansion}. A summary of our algorithm in the \acrshort{ll} setting is provided in Algorithm~\ref{alg:algo}. Our implementation is available at \url{https://github.com/KingSpencer/DBULL}. The implementation details are provided in Appendix C. We explain the contribution of the sufficient statistics to the probabilistic density function of our problem and knowledge
preservation in Section~\ref{sec:ss}. 
\subsection{Variational Inference and ELBO Derivation}\label{sec:label}
In practice, it is often infeasible to obtain the exact posterior distribution since the normalizing constant in the posterior distribution is intractable. \acrfull{mcmc} methods are a family of algorithms that provide a systematic way to sample from the posterior distribution but is often slow in a high-dimensional parameter space.  Thus, effective alternative methods are needed. 
Variational inference is a promising alternative, which approximates the posterior distribution by casting inference as an optimization problem. It aims to find a surrogate distribution that is the most similar to the distribution of interest over a class of tractable distributions  that can minimize the Kullback-Leibler (KL) divergence to the exact posterior distribution. Minimizing the KL divergence between $q(\bm{y}, \bm{z}, \bm{\phi}, \bm{v}|\bm{x})$ and $p(\bm{y}, \bm{z}, \bm{\phi}, \bm{v}|\bm{x})$ in our setting is equivalent to maximizing the Evidence Lower Bound (ELBO), where $q(\bm{y}, \bm{z}, \bm{\phi}, \bm{v}|\bm{x})$ is the variational posterior distribution used to approximate the true posterior distribution. To make it easier for the readers to understand the core idea, we provide a high-level explanation of variational inference  and mathematical details can be found in Appendix A. 

Given the generative process in Section~\ref{sec:gp} and using Jensen's inequality,
\begin{align}
\log p(\bm{x}) &\geqslant \mathbb{E}_{q(\bm{y}, \bm{z}, \bm{\phi}, \bm{v}|\bm{x})}\left\{
    \log\frac{p(\bm{x}, \bm{y}, \bm{z}, \bm{\phi}, \bm{v})}{q(\bm{y}, \bm{z}, \bm{\phi}, \bm{v}|\bm{x})}\right\}\\
    &=\mathcal{L}_{\textrm{ELBO}}(\bm{x}),\notag
\end{align}
For simplicity, we assume that $q(\bm{y}, \bm{z}, \bm{\phi}, \bm{v}|\bm{x})=q_{\psi}(\bm{z}|\bm{x})q(\bm{y})q(\bm{v})q(\bm{\phi})$. Thus, the ELBO is 
\begin{align} 
            \begin{split}
&\mathbb{E}_{q(\bm{y}, \bm{z}, \bm{\phi}, \bm{v}|\bm{x})}
    \left[\log\frac{p_\theta(\bm{x}|\bm{z})p(\bm{z}|\bm{y}, \bm{\phi})p(\bm{y}|\bm{v})p(\bm{v})p(\bm{\phi})}{q_{\psi}(\bm{z}|\bm{x})q(\bm{y})q(\bm{v})q(\bm{\phi}))}\right]
    \\
    &= \mathbb{E}_{q(\bm{y}, \bm{z}, \bm{\phi}, \bm{v}|\bm{x})}
    \left[\log p_\theta(\bm{x}|\bm{z})\right]
    +\mathbb{E}_{q(\bm{y}, \bm{z}, \bm{\phi}, \bm{v}|\bm{x})}
    \left[\log p(\bm{z}|\bm{y}, \bm{\phi})\right]\\
    &- \mathbb{E}_{q(\bm{y}, \bm{z}, \bm{\phi}, \bm{v}|\bm{x})}
    \left[\log q_{\psi}(\bm{z}|\bm{x})\right] 
    \\
   &+ \mathbb{E}_{q(\bm{y}, \bm{z}, \bm{\phi}, \bm{v}|\bm{x})}
    \left[\log p(\bm{y}|\bm{v})\right] 
    + \mathbb{E}_{q(\bm{y}, \bm{z}, \bm{\phi}, \bm{v}|\bm{x})}
    \left[\log p(\bm{v})\right] 
    \\
    &- \mathbb{E}_{q(\bm{y}, \bm{z}, \bm{\phi}, \bm{v}|\bm{x})}
    \left[\log q(\bm{y})\right] 
    - \mathbb{E}_{q(\bm{y}, \bm{z}, \bm{\phi}, \bm{v}|\bm{x})}
    \left[\log q(\bm{v})\right] 
    \\
    &- \mathbb{E}_{q(\bm{y}, \bm{z}, \bm{\phi}, \bm{v}|\bm{x})}
    \left[\log q(\bm{\phi})\right] 
    + \mathbb{E}_{q(\bm{y}, \bm{z}, \bm{\phi}, \bm{v}|\bm{x})}
    \left[\log p(\bm{\phi})\right] 
        \end{split}
\label{eq:all_elbo}
\end{align}
We assume our variational distribution takes the form of 

\begin{align} 
     \begin{split}
&q(\bm{y}, \bm{z}, \bm{\phi}, \bm{v}|\bm{x})=q_{\psi}(\bm{z}|\bm{x})q(\bm{\phi})q(\bm{v})q(\bm{y})   \\
   &=\mathcal{N}(\bm{\mu}(\bm{x};\psi), \textrm{diag}(\bm{\sigma}^2(\bm{x}; \psi)))\prod_{t=1}^{T-1} q_{\eta_t}(v_t)\prod_{t=1}^T q_{\zeta_t}(\bm{\phi}_t)\prod_{n=1}^{N}q_{\rho_n}(\bm{y}_n)\\
    &=\mathcal{N}(\bm{\mu}(\bm{x};\psi), \textrm{diag}(\bm{\sigma}^2(\bm{x}; \psi))) \prod_{t=1}^{T-1}\textrm{Beta}(\eta_{t_1}, \eta_{t_2})\\
    &\prod_{t=1}^T\mathcal{N}(\bm{\mu}_t|\bm{m}_t, (\bm{\beta}_t\bm{\Lambda}_t)^{-1})\mathcal{W}(\bm{\Lambda}_t|W_t, \nu_t)\prod_{n=1}^{N} \textrm{Mult}(T, \rho_n),
    \end{split}
\label{eq:variational_new}
\end{align}
where we denote $f_\psi(\bm{x})=(\bm{\mu}(\bm{x};\psi), \bm{\sigma}^2(\bm{x}; \psi))$, which is a neural network parameterized by $\psi$, $T$ is the number of mixture components in the DP of the variational distribution, $\bm{z}_n\sim \mathcal{N}(\bm{z}_n|\bm{\mu}_t, \bm{\Lambda}_t^{-1})$, $\bm{\phi}_t=(\bm{\mu}_t, \bm{\Lambda}_t)$, and $\textrm{Mult}(T, \rho_n)$ is a Multinomial distribution. The notation definitions in equation~\eqref{eq:all_elbo}, \eqref{eq:variational_new} and  \eqref{eq:elbo_derivation} are provided in Table~\ref{tab:notation}. Our inference strategy starts with only one mixture component  and uses \acrshort{cerr} technique described in Section~\ref{sec:expansion} to either increase or merge the number of clusters.


\subsection{General Parameter Learning Strategy }\label{sec:learning}
In equation~\eqref{eq:all_elbo}, there are mainly two types of parameters which we need to optimize. The first type includes parameters $\theta$ and $\psi$ in the neural network. The other type involves the latent cluster membership $\bm{y}$ and the parameters for the \acrshort{dp} Gaussian mixture model. 

In order to perform joint inference for both types of parameters, we adopt the alternating  optimization strategy. First we update the neural network parameters ($\theta$ and $\psi$) to learn the latent representation $\bm{z}$ given the DP Gaussian mixture parameters. This is achieved by optimizing $\mathcal{L}_{\textrm{ELBO-VAE}}$, which only involves the first three terms of equation~\eqref{eq:all_elbo} that make a contribution to optimize $\theta$, $\psi$ and $\bm{z}$. Under our variational distribution assumptions in \eqref{eq:variational_new}, by taking advantage of the reparameterization trick \citep{kingma2014auto} and  the Monte Carlo estimate of expectations, we obtain
\begin{align}
\begin{split}
&\mathcal{L}_{\textrm{ELBO}-\textrm{VAE}}(\bm{x}) \\
=& -\frac{1}{2} \sum_{k=1}^{T} N_k \nu_k \left\{ \textrm{Trace}(\bm{U}_k W_k)\right\}-\frac{1}{2} \sum_{k=1}^{T} N_k \nu_k\left\{(\bar{z}_k-\bm{m}_k)^T W_k (\bar{z}_k - \bm{m}_k)\right\} \\
&-\frac{1}{2}\frac{1}{L}\sum_{l=1}^L\sum_{i=1}^N\sum_{j=1}^D\bigg(\log(\bm{\sigma}(\bm{z}, \theta)^2)_j^{(l)}+ \frac{
    \left(\bm{x}_{ij}- \bm{\mu}(\bm{z};\theta)_j^{(l)}\right)^2}{(\bm{\sigma}(\bm{z};\theta)^2)_j^{(l)}}\bigg) +\frac{1}{2}\log(\textrm{Det}(2\pi e \Sigma)).
    \end{split}
\label{eq:elbo_derivation}
\end{align}

\begin{table}[t]
  \caption{Notations in $\mathcal{L}_{\textrm{ELBO}-\textrm{VAE}}(\bm{x})$.} 
   \label{tab:notation}
   \small 
   \centering 
   \begin{tabular}{l} 
   \toprule[\heavyrulewidth]\toprule[\heavyrulewidth]
   \textbf{Notations in the ELBO} \\ 
   \midrule
   $\theta$: parameters in the decoder.\\
   $\psi$: parameters in the encoder.\\
   $N$: the total number of observations.\\
   $T$: the total number of clusters.\\
   $D$: the dimension of the latent representation $\bm{z}$.\\
   $\Sigma: \textrm{diag}(\bm{\sigma}^2(\bm{x};\psi))$. \\
   $\bm{x}_{ij}$: the $j$th dimension of the $n$th observation. \\
   $\bm{y}_n$: cluster membership for the $n$th observation.\\ 
   $p(\bm{y}_n=k) = \gamma_{ik}, N_k = \sum_{n=1}^N \gamma_{nk}$.  \\
   $L$: the number of Monte Carlo samples in Stochastic\\ Gradient
   Variational Bayes (SGVB).\\ 
   $\hat{\bm{z}}_n=\frac{1}{L}\sum_{l=1}^L \bm{z}_n^{(l)}.$ 
   $\bar{\bm{z}}_k=\frac{1}{N_k}\sum_{n=1}^{N}\gamma_{nk}\hat{\bm{z}}_n$.\\
     $\bm{U}_k = \frac{1}{N_k}\sum_{n=1}^N \gamma_{nk} (\hat{\bm{z}}_n - \bar{\bm{z}}_k)(\hat{\bm{z}}_n - \bar{\bm{z}}_k)^T.$ \\
     $\beta_k = \beta_0+ N_k$: the posterior scalar precision in NW distribution. \\
    $\bm{m}_k=\frac{1}{\beta_k}(\beta_0\bm{m}_0+N_k\bar{z}_k)$: the posterior mean of cluster $k$. \\
    $W_k^{-1} = W_0^{-1} + N_k S_k + \frac{\beta_0 N_k}{\beta_0 + N_k}(\bar{\bm{z}}_k- \bm{m}_0)(\bar{\bm{z}}_k- \bm{m}_0)^T$.\\ 
    $\nu_k = \nu_0+N_k$: the $k$th posterior degrees of freedom of NW. \\
   \bottomrule[\heavyrulewidth] 
   \end{tabular}
   \label{tab:notation}
\end{table}

We provide the notations in Table~\ref{tab:notation}. The derivation details are provided in Appendix A.  
Then, we update the \acrshort{dp} Gaussian mixture parameters and the cluster membership given the current neural network parameters $\theta$, $\psi$ and the latent representation $\bm{z}$. This allows us to use improved latent  representation to infer latent cluster memberships and the updated clustering will in turn facilitate learning latent knowledge representation. The update equations for \acrshort{dp} mixture model parameters can be found in \cite{blei2006variational}. 
We describe the core idea of automatic \acrshort{cerr} in our inference in Section~\ref{sec:expansion} to explain how we start with only one cluster and achieve dynamic model expansion by creating new mixture components (clusters) given new data in \acrshort{ll}. 

Our general parameter learning strategy via variational inference may seem straightforward for a batch setting at first glance. However, both the derivation and the implementation is nontrivial especially when incorporating our new components in the end-to-end inference procedure to address the additional challenges in a \acrshort{ll} setting. For illustration purposes, we choose to describe the high level core idea of our inference procedure. The main difficulty lies in how to adapt our inference algorithm from a batch setting to a \acrshort{ll} setting, which requires us to overcome catastrophic forgetting, maintain past knowledge and develop a dynamic model that can expand with automatic cluster discovery and redundancy removal capacity. Next, we describe our novel solutions.

\subsection{Our Ingredients for Alleviating Catastrophic Forgetting}\label{sec:forgetting}

\emph{Catastrophic forgetting} or \emph{catastrophic interference} is a dramatic issue for \acrshort{dnn} as witnessed in \acrshort{sll}  \citep{kirkpatrick2017overcoming, shin2017continual}.
In our \acrshort{ull} setting, the issue is even more challenging since we have more sources than  \acrshort{sll} that may lead to abrupt model performance decrease due to the interference of training with new data. The first source is the same as in  \acrshort{sll}  when the \acrshort{dnn} forget previously learned information upon learning new information. Additionally, in an unsupervised setting, the model is not able to recover the learned cluster membership and clustering related parameters in the \acrshort{dp} mixture model when the previous data is no longer available, or when the previous learned information of \acrshort{dnn} has been wiped out upon learning new information, since the clustering structure learned depends on the latent representation of the raw data, which is determined by the \acrshort{dnn}' parameters and the data distributions. 

To resolve these issues, we develop our own novel solution via a combination of two ingredients: (1) generating and replaying a fixed small number of samples based on our generative process in Section~\ref{sec:gp} given the current \acrshort{dnn} and \acrshort{dp} Gaussian mixture parameter estimates, which is a computationally effective byproduct of our algorithm; and, (2) developing a novel hierarchical sufficient statistics knowledge preservation strategy to remember the clustering information in an unsupervised setting.

We choose to replay a number of generative samples to preserve the previous data distribution instead of using a subset of past real data, since storing past data may require large memory and such data storage and replay may not be feasible in real big data applications. More details of replaying deep generative samples over real data in \acrshort{ll} have been discussed in \cite{shin2017continual}. Moreover, our proposal to use sufficient statistics is novel and has the advantage of allowing incremental updates of the clustering information as new data arrive without the need of access to previous data because of the additive property of sufficient statistics. We introduce this novel strategy in the next section.

\subsection{Sufficient Statistics for Knowledge Preservation}\label{sec:suff}
As \acrshort{ll} is an emerging field, and there is no well-accepted knowledge definition or an appropriate representation scheme to efficiently maintain past knowledge from seen data. Researchers have adopted prior distributions \citep{nguyen2017variational} or model parameters \citep{ lee2019learning} to represent past knowledge in most \acrshort{sll} problems, where achieving high prediction accuracy incrementally is the main objective. However, there is no guidance on preserving past knowledge in an unsupervised learning setup.

We propose a novel knowledge preservation strategy in \acrshort{dbull}. In our problem, there are  two types of knowledge to maintain. The first contains previously learned \acrshort{dnn}' parameters needed to encode the  latent knowledge representation $\bm{z}$ of the raw data and the reconstruction of the real data from $\bm{z}$. The other involves the \acrshort{dp} Gaussian mixture parameters to represent different cluster characteristics and different cluster mixing proportions. Our novel knowledge representation scheme uses \emph{hierarchical sufficient statistics} to preserve the information related to the \acrshort{dp} Gaussian mixture. We develop a sequential updating rule to update our knowledge.

Assume that we have encountered $N$ datasets $\{\mathcal{D}_j\}_{j=1}^{N}$ and each time only one dataset $\mathcal{D}_j$ can be in memory. While in memory, each dataset can be divided into $M$ mini-batches $\{\mathcal{B}_i\}_{i=1}^M$. To define the sufficient statistics, we first define the \emph{global} parameters of the \acrshort{dp} Gaussian mixture as probabilities of each mixture component (cluster) $\pi_{k}s$ and the mixture parameters $(\bm{\mu}_k^*, \bm{\sigma}_k^*)$ for each cluster $k$. We define the \emph{local} parameters as the cluster membership for each observation in memory. To remember the characteristics of all encountered data and the local information of the current dataset, we memorize three levels of sufficient statistics. 
The $i$th {\em mini-batch sufficient statistics} $S_k^{j,i}=(N_k(\mathcal{B}_i), s_k(\mathcal{B}_i))$ of the current dataset $D_j$, where $s_k(\mathcal{B}_i)=\sum_{n\in\mathcal{B}_i}\hat{\gamma}_{nk}t(z_n)$ and $t(\bm{z}_n)$ is the sufficient statistics to represent a distribution within the exponential family (Gaussian distribution is within the exponential family and $t(\bm{z}_n) = (\bm{z}_n, \bm{z}_n^T\bm{z}_n)$ in our case) and $\hat{\gamma}_{nk}$ represents the estimated probability of the $n$th observations in mini-batch $\mathcal{B}_i$ belonging to cluster $k$. We also define the {\em stream sufficient statistics} $S_k^j=\sum_{i=1}^M
S_k^{j,i}$ of dataset $\mathcal{D}_j$ and the {\em overall sufficient statistics}  $S_k^0 = (N_k, s_k(\bm{z}))$ of all encountered datasets $\{\mathcal{D}_j\}_{j=1}^{N}$.

To efficiently maintain and update our knowledge, we develop our updating algorithm as: (1) substract the old summary of each mini-batch and update the local parameters; (2) compute a new summary for each mini-batch; and, (3) update the stream sufficient statistics for each cluster learned in the current dataset.
\begin{align}
\label{eq:j_batch}
&S_k^j \leftarrow  S_k^j - S_k^{j,i},   \\
\label{eq:ji_batch}
&S_k^{j, i} \leftarrow\left(\sum_{n\in\mathcal{B}_i} \hat{\gamma}_{nk}, \sum_{n\in\mathcal{B}_i} \hat{\gamma}_{nk} t(\bm{z}_n) \right), \\
\label{eq:i_batch}
&S_k^j \leftarrow  S_k^j + S_k^{j,i}.  
\end{align}

For the dataset $\mathcal{D}_j$ in the learning phase, we repeat the updating process multiple iterations to refine our training while learning the \acrshort{dp} Gaussian mixture parameters and the cluster membership. Finally, we update the overall sufficient statistics by
$ S_k^0\leftarrow S_k^0 + S_k^j$.
The correctness of the algorithm is guaranteed by the additive property of the sufficient statistics. 

\subsection{Contribution of Sufficient Statistics to 
Alleviate Forgetting}\label{sec:ss}
The sufficient statistics alleviate forgetting by preserving data characteristics and allow sequential updates in \acrshort{ll} without the need of saving real data. To be precise, the sufficient statistics allow us to update the log-likelihood and the ELBO sequentially since both terms are linear functions of the expectation of the sufficient statistics. Given the expected sufficient statistics, we are able to evaluate 
the first two terms of  $\mathcal{L}_{\textrm{ELBO}-\textrm{VAE}}(\bm{x})$ in equation~\eqref{eq:elbo_derivation} and $p(\bm{z}\vert \bm{y})p(\bm{y}|\bm{v})$ in the joint probability density function of our model  in  equation~\eqref{eq:pdf}. Next, we provide mathematical derivations to illustrate this. 

Define sufficient statistics of all data $S^0=(S_1^0, S_2^0, …., S_k^0)$, where $k$ is the number of clusters.  Define $S_k^0=(N_k, \sum_{n=1}^N r_{nk}t(\bm{z}_n))$, where $t(\bm{z}_n) = (\bm{z}_n, \bm{z}_n^T \bm{z}_n)$ and $r_{nk}$ denotes the probability of the $n$th observation belonging to cluster $k$, $N_k=\sum_{n=1}^N r_{nk}$, and $N$ is the total number of observations. Given the sufficient statistics and current mixture parameters $(\bm{\mu}_k^*, \bm{\sigma}_k^*)$ and $\pi(v)$, we can evaluate $\sum_{k=1}^\infty N(\bm{z}_n|\bm{\mu}_k^*, \bm{\sigma}_k^{*2} I)P(\bm{y}_n=k|\pi(v))$ in the joint probability density function in equation~\eqref{eq:pdf} without storing each latent representation $\bm{z}_n$ for all data. Similarly, we can also evaluate the first two terms of  $\mathcal{L}_{\textrm{ELBO}-\textrm{VAE}}(\bm{x})$ in equation~\eqref{eq:elbo_derivation} with notations defined in Table~\ref{tab:notation}.

\subsection{Cluster Expansion and Redundancy Removal Strategy}\label{sec:expansion}
Our model starts with one cluster and, as we have new data with different characteristics, we expect the model to either dynamically  grow the number of clusters or merge clusters if they have similar characteristics. To achieve this, we perform birth and merge moves in a similar fashion as the Nonparametric Bayesian literature \citep{hughes2013memoized} to allow automatic \acrshort{cerr}. However, we would like to emphasize that our work is different from \cite{hughes2013memoized} since our merge moves have extra constraints. To avoid losing information about clusters learned earlier, we only allow merge moves between two novel clusters from the birth move or one existing cluster with a newborn cluster. Two previously existing clusters cannot be merged. Reference \cite{hughes2013memoized}  is designed for a batch learning, thus, it does not require this constraint (but in \acrshort{ll} this constraint is important for avoiding information loss). 

It is challenging to give birth to new clusters with streaming data since the number of observations may not be sufficient to inform good proposals. To resolve this issue, we follow \cite{hughes2013memoized} by collecting a subsample of data for each learned cluster $k$.
Then, we cache the samples in the subsample if the probability of the $n$th observation to be assigned to cluster $k$ is bigger than a threshold of value 0.1. This value has been suggested by \cite{hughes2013memoized}. In this paper, we try to choose commonly used parameters in the literature and avoid dataset specific tuning as much as possible. We fit the \acrshort{dp} Gaussian mixture to the cached samples with one cluster and expand the  model with 10 novel clusters. However, only adopting the birth moves may overcluster the observations into different clusters. After the birth move, we merge the clusters by (1) selecting candidate clusters to merge and by (2) merging two selected clusters if ELBO improves. The candidate clusters are selected if the marginal likelihood of two merged clusters is bigger than the marginal likelihood when keeping the two clusters separate. 

\begin{algorithm}[p]
\caption{Variational Inference for DBULL}
\label{alg:algo}
\begin{algorithmic}[1]
{
\small
\STATE \textbf{Initialization:} \\
Initialize the \acrshort{dnn}, variational distributions and the hyperparameters for \acrfull{dpmm}. \\
\FOR{$j=1, 2, \ldots, N $ datasets in memory}
\FOR{$\textrm{epoch} = 1, 2, \ldots$}
\FOR{$h=1, 2, \ldots, H $ iterations}
\STATE Update the weights $\omega=\{\psi, \theta\}$ of the encoder and decoder via $\omega^{(h+1)}=\omega^{(h)} + \eta\frac{\partial L_{\textrm{ELBO-VAE}}(\bm{x})}{\partial \omega^{(h)}}$ to maximize $\mathcal{L}_{\textrm{ELBO}-\textrm{VAE}}(\bm{x}) $ in equation~\ref{eq:elbo_derivation} given current \acrshort{dpmm} parameters, where $\eta$ is the learning rate when using stochastic gradient descent. 
\ENDFOR
\STATE Compute the deep representation $\bm{z}$ of observations $\bm{x}$ using the encoder
$\bm{z}=f_\psi(\bm{x})=(\bm{\mu}(\bm{x};\psi), \bm{\sigma}^2(\bm{x}; \psi))$.
\FOR{$i=1, 2, \ldots, M_j$ mini-batch of dataset $j$}
\WHILE{The ELBO of the \acrshort{dpmm} has not converged}
\STATE Visit the $i$th mini-batch of $\bm{z}$ in a full pass  in the $j$th dataset.  
\STATE Update the \emph{mini-batch sufficient statistics} and \emph{stream sufficient statistics} in Section~\ref{sec:suff} via equation~\ref{eq:j_batch} and \ref{eq:ji_batch}.
\STATE  Update the local and global parameters of \acrshort{dpmm} using mini-batch $i$ of dataset $j$ via standard variational inference for \acrshort{dpmm} \cite{blei2006variational} provided in Appendix A. 
\STATE Perform \emph{cluster expansion} described in Section~\ref{sec:expansion} to propose new clusters. 
\STATE Perform \emph{redundancy removal} described in Section~\ref{sec:expansion} to merge clusters if the ELBO improves.
\ENDWHILE
\ENDFOR
\STATE Update the \emph{overall sufficient statistics} defined in Section~\ref{sec:suff} via equation~\ref{eq:i_batch}.
\ENDFOR
\ENDFOR
}
\STATE \textbf{Output}: Learned \acrshort{dnn}, variational approximation to the posterior, deep latent representation, cluster assignment for each observation, and cluster representative in the latent space. 
\end{algorithmic}
\end{algorithm}
\section{Experiments}\label{sec:experiment}
\textbf{Datasets.}
We adopt the most common text and image benchmark datasets in \acrshort{ll} to evaluate the performance of our method. The \emph{MNIST} database of 70,000 handwritten digit images \citep{lecun1998gradient} is widely used to evaluate deep generative models \citep{kingma2014auto, xie2016unsupervised,johnson2016composing, goyal2017nonparametric, jiang2017variational}  for representation learning and \acrshort{ll} models in both supervised \citep{kirkpatrick2017overcoming, nguyen2017variational, shin2017continual} and unsupervised learning contexts \citep{rao2019continual}. 
To provide a fair comparison with state-of-the-art competing methods and easy interpretation, we mainly use \emph{MNIST} to evaluate the performance of our method and interpret our results with intuitive visualization patterns. To examine our method on more complex datasets, we use text \emph{Reuters10k} \citep{lewis2004rcv1} and image \emph{STL-10} \citep{coates2011analysis} databases. STL-10 is at least as hard as a well-known image database CIFAR-10 \citep{krizhevsky2009learning} since STL-10 has fewer labeled training examples within each class. The summary statistics for the datasets are provided in Table~\ref{tb:DATASET}. 

\begin{table}[t]
\caption{Summary statistics for benchmark datasets.}
\label{tb:DATASET}
\scriptsize
\centering
\begin{tabular}{lcccc}  \toprule
Dataset & \# Samples & Dimension & \# Classes \\
\midrule
 MNIST & 70000 & 784 & 10 \\ 

 Reuters10k & 10000 & 2000 & 4 \\

 STL-10 & 13000 & 2048 & 10 \\ 
 \bottomrule 
\end{tabular}
\end{table}
We adopt the same neural network architecture as in  \cite{jiang2017variational}. All values of the tuning parameters and the implementation details in the \acrshort{dnn} are provided in \ref{app:implementation}. Our implementation is publicly available at \url{https://github.com/KingSpencer/DBULL}. 

\textbf{Competing Methods in Unsupervised Lifelong Learning.} 
\acrshort{curl} is the only lifelong unsupervised learning method currently with both representation learning and new cluster discovery capacity, which makes \acrshort{curl} the latest, most related, and comparable method to ours. We use \textbf{CURL-D} to represent \acrshort{curl}  without the true number of clusters provided but to detect it \textbf{D}ynamically given unlabelled streaming data. 

\textbf{Competing Methods in a Classic Batch Setting.}
Although designed for \acrshort{ll}, to show the generality of \acrshort{dbull} in a batch training mode, we compare it against recent deep (generative) methods with representation learning and clustering capacity designed for batch settings, including DEC \citep{xie2016unsupervised}, VaDE 
\citep{jiang2017variational}, CURL-F \citep{rao2019continual} and VAE+DP.  CURL-F represents \acrshort{curl} with the true number of clusters provided as a \textbf{F}ixed value. VAE+DP fits a \acrfull{vae} \cite{kingma2014auto} to learn latent representations first and then uses a \acrshort{dp} to learn clustering in two separate steps.
We list the  capabilities of different methods in Table~\ref{tb:diff}. 

\begin{table}[htbp]
\scriptsize
\centering
\setlength{\tabcolsep}{2pt}
\caption{DBULL and competing methods capacity comparison.}
\begin{tabular}{lccrrrr}
\toprule
&\multicolumn{2}{c}{\textbf{Lifelong Learning}} & \multicolumn{4}{c}{\textbf{Batch Setting} }\\
\cmidrule(r){2-3}\cmidrule(l){4-7}
        & 
        \textbf{DBULL} &
        CURL-D&
        DEC &
        VaDE &
        VAE+DP &
        CURL-F  
         \\
\midrule
    Representation Learning & 
        \textbf{yes} &
        \textbf{yes} &
        \textbf{yes} &
        \textbf{yes} &
        \textbf{yes} &
        \textbf{yes}
        \\
   Learns \# of Clusters &
        \textbf{yes} &
        \textbf{yes} &
        no &
        no &
        \textbf{yes} &
        no 
        \\ 
        
    Dynamic Expansion &
        \textbf{yes} &
        \textbf{yes} &
        no &
        no &
        \textbf{yes} &
        no 
        \\
    Overcome Forgetting &
        \textbf{yes} &
        \textbf{yes}&
        no &
        no &
        no &
        \textbf{yes} 
        \\
    
\bottomrule
\end{tabular}

\label{tb:diff}
\end{table}

\textbf{Evaluation Metrics.}
One of the main objectives of our method is to perform new cluster discovery with streaming non-stationary data. Thus, it is desired if our method can achieve superior clustering quality in both \acrshort{ull} and batch settings. We adopt the clustering quality metrics including Normalized Mutual Information (NMI), Adjusted Rand Index (ARI), Homogeneity Score (HS), Completeness Score (CS) and V-measure Score (VM). These are all normalized metrics ranging from zero to one, and larger values indicate better clustering quality. NMI, ARI and VM of value one represent perfect clustering as the ground truth. CS is a symmetrical metric to HS. Detailed definitions for these metrics can be found in \cite{rosenberg2007v}.



\subsection{Lifelong Learning Performance Comparison}
\textbf{Experiment Objective.}
It is desired if \acrshort{ll} methods can adapt a learned model to new data while retaining the information learned earlier. The objective of this experiment is to demonstrate  \acrshort{dbull} has such desired capacity and effectiveness compared with state-of-the-art \acrshort{ll} methods such that there is no dramatic performance decrease on past data even if the model has been updated with new information. 

\textbf{Experiment Setup.} To evaluate the performance of \acrshort{dbull}, we
adopt the most common experiment setup called \emph{Split MNIST} in \acrshort{ll}, which used images from MNIST \citep{zenke2017continual, nguyen2017variational}. We divide MNIST into 5 disjoint subsets with each subset containing 10,000 random samples of two digit classes in the order of digits 0-1, 2-3, 4-5, 6-7 and 8-9, denoted as $\text{DS}_1$, $\text{DS}_2$, $\text{DS}_3$, $\text{DS}_4$, and $\text{DS}_5$. Each dataset is divided into 20 subsets that arrive in a sequential order to mimic a \acrshort{ll} setting. We denote $\text{DS}_{i:j}$ as all data from $\text{DS}_i$ to $\text{DS}_j$, where $i, j=1, 2, \ldots, 5$.

\begin{figure}[t]
\centerline{\includegraphics[scale=0.3]{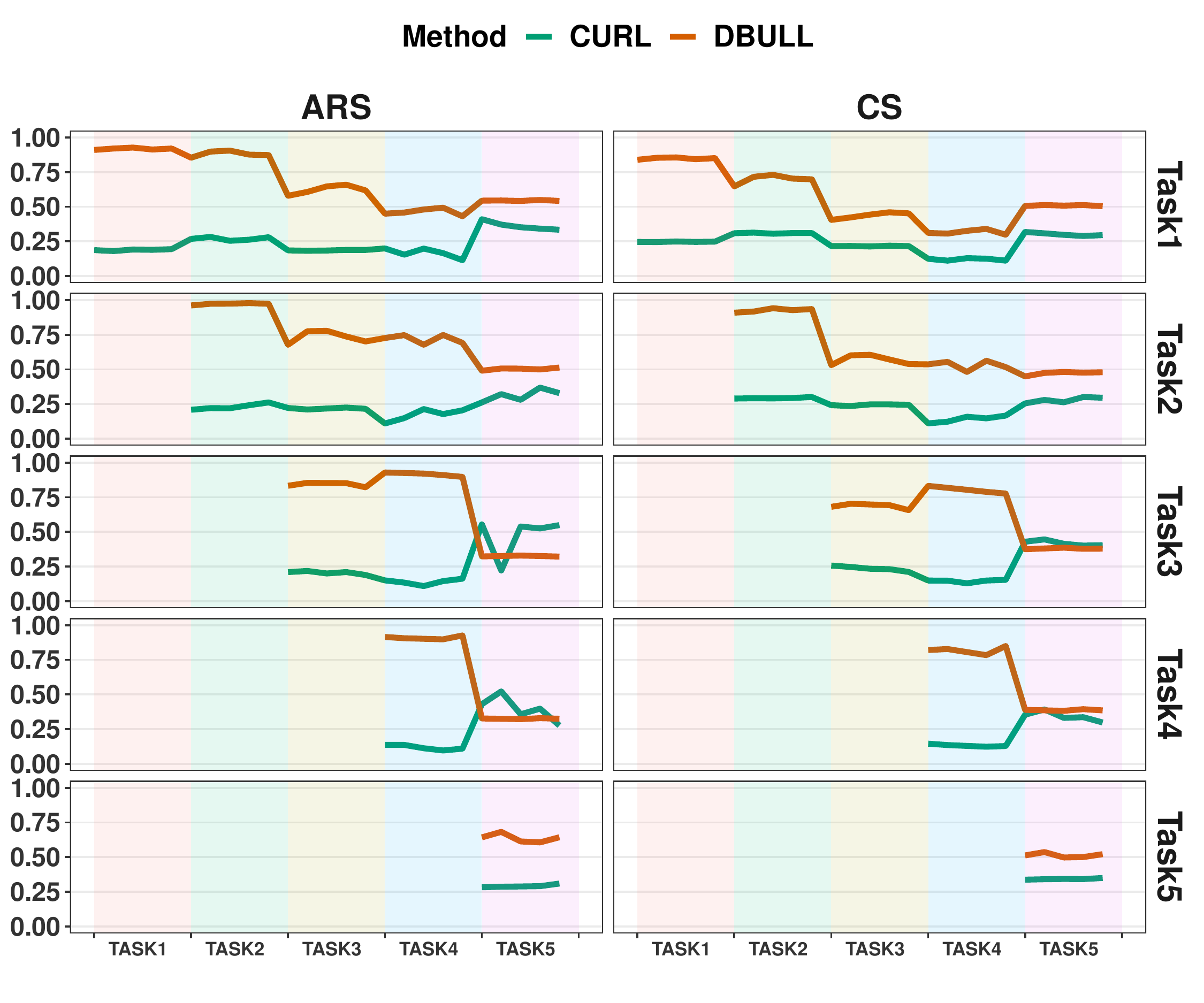}}
\caption{ Clustering quality performances in terms of ARS and CS measured on each task after
sequential training from $\textrm{TASK}_1$ to $\textrm{TASK}_5$ across every 500 iterations for each task.}
\label{fig:ARS}
\end{figure}

\begin{figure}[H]
\centerline{\includegraphics[scale=0.3]{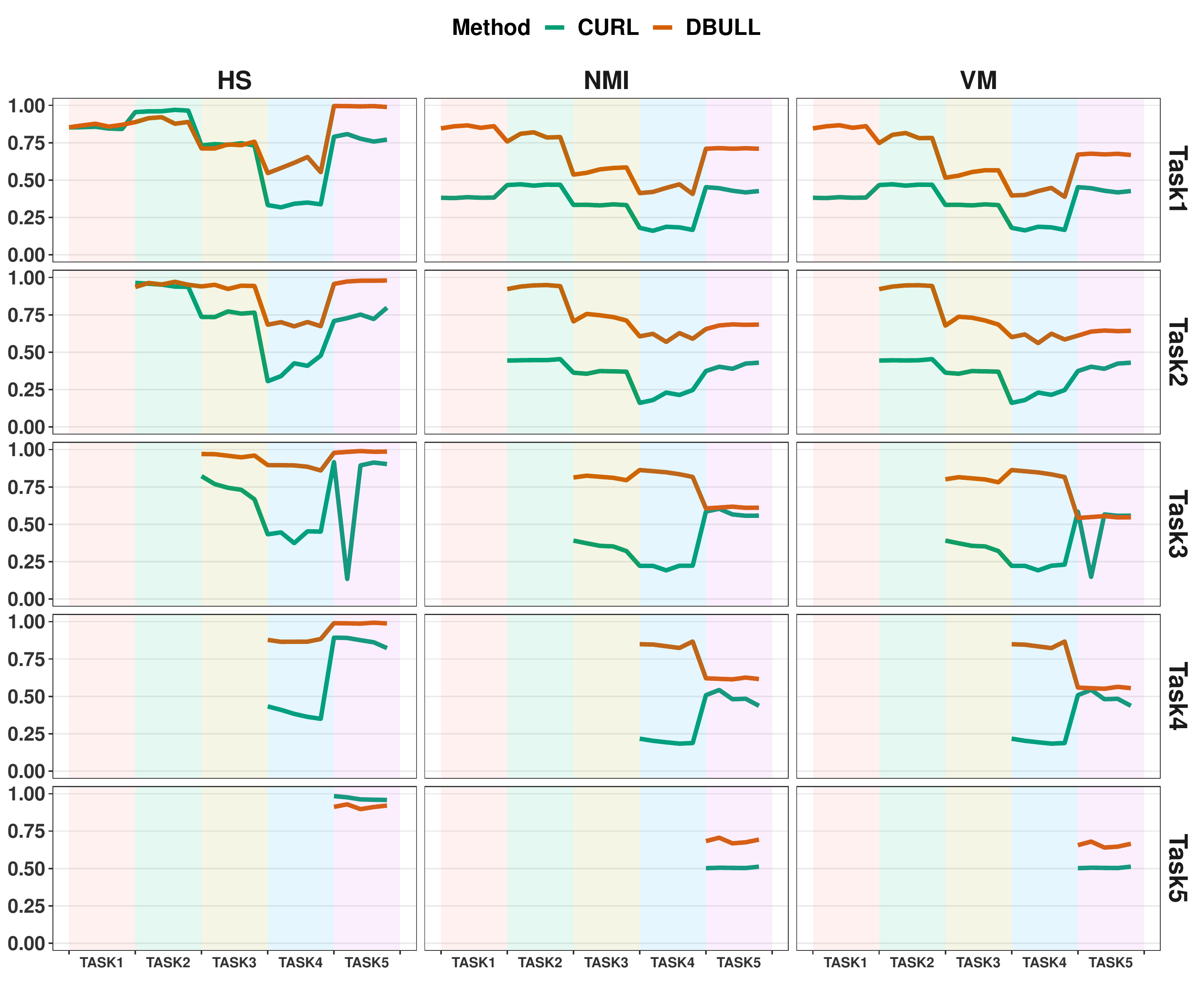}}
\caption{Clustering quality performances in terms of HS, NMI and VM measured on each task after
sequential training from $\textrm{TASK}_1$ to $\textrm{TASK}_5$ across every 500 iterations for each task.}
\label{fig:HS}
\end{figure}

\textbf{Discussion on Performance.}
To check if
our method has dramatic performance loss due to catastrophic forgetting, we sequentially train our method \acrshort{dbull} and its \acrshort{ll} competitor CURL-D on $\textrm{DS}_1, \textrm{DS}_2, \ldots, \textrm{DS}_5$. We define \emph{$\textrm{TASK}_i$} as training on \emph{$\textrm{DS}_i$}, where $i=1, 2, \ldots, 5$. We measure the performance of $\textrm{TASK}_1$  after training $\textrm{TASK}_1$, $\textrm{TASK}_2$, $\textrm{TASK}_3$, $\textrm{TASK}_4$, $\textrm{TASK}_5$ with datasets $\textrm{DS}_1, \textrm{DS}_{1:2}, \textrm{DS}_{1:3}, \textrm{DS}_{1:4}$, and $\textrm{DS}_{1:5}$, the performance of $\text{TASK}_2$ after training $\textrm{TASK}_2, \textrm{TASK}_3, \textrm{TASK}_4, \textrm{TASK}_5$ with datasets $\textrm{DS}_2, \textrm{DS}_{2:3}, \textrm{DS}_{2:4}, \textrm{DS}_{2:5}$, etc. We report the \acrshort{ll} clustering quality performance for each task after sequential training five tasks in Fig.~\ref{fig:ARS} and Fig.~\ref{fig:HS}. 

Fig.~\ref{fig:ARS} and Fig.~\ref{fig:HS} reflect that \acrshort{dbull} has
better performance in handling catastrophic forgetting than
CURL-D since \acrshort{dbull} has slightly less performance drop than CURL-D for previous tasks in almost all scenarios in terms of nearly all clustering metrics. 

\begin{figure}[H]
\centerline{\includegraphics[scale=0.3]{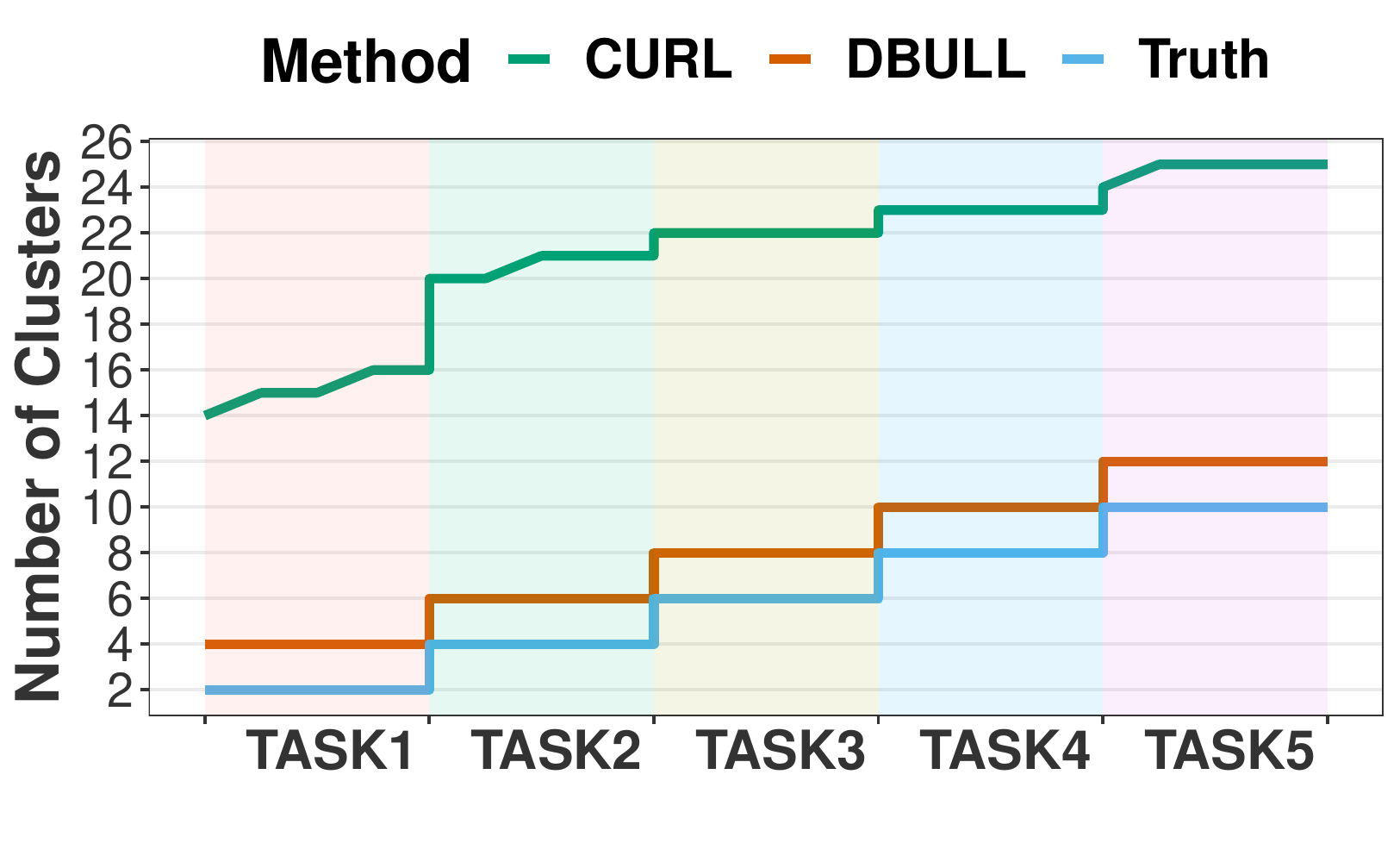}}
\caption{Number of clusters detected by CURL-D, DBULL after sequentially training from $\textrm{TASK}_1$ to $\textrm{TASK}_5$ across every 500 iterations for each task compared with the ground truth.}
\label{fig:cluster}
\end{figure}

\begin{figure}[H]
\centerline{\includegraphics[scale=0.42]{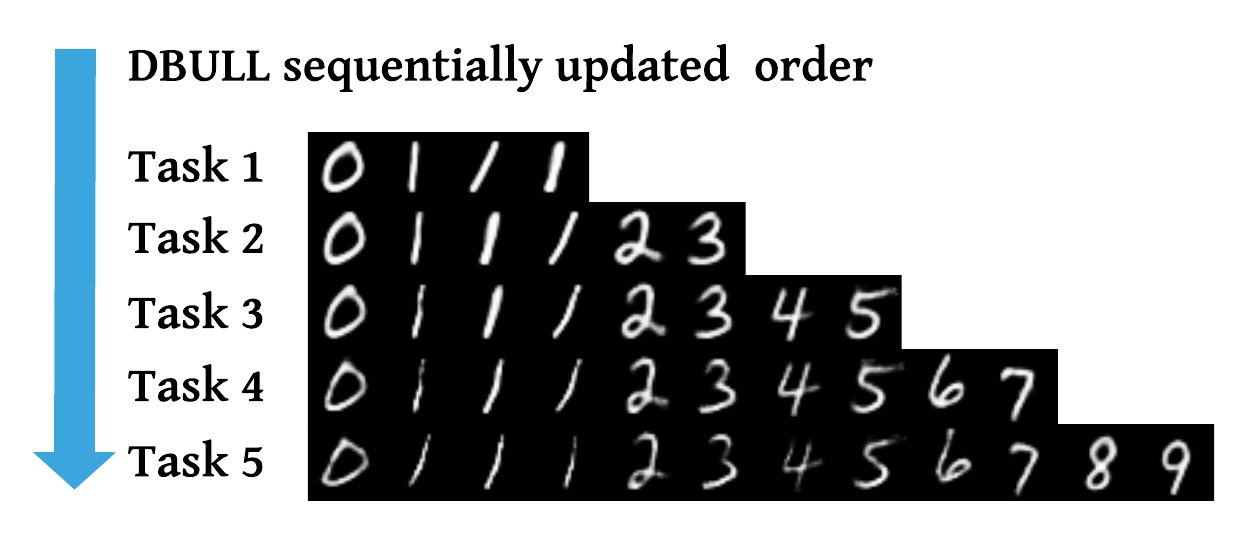}}
\caption{Decoded images using the DP Gaussian mixture  posterior mean after sequentially trained from $\textrm{TASK}_1$ to $\textrm{TASK}_5$ using DBULL.}
\label{fig:cluster_mean}
\end{figure}

Fig.~\ref{fig:cluster} reflects that \acrshort{dbull} has advantages over CURL-D in handling overclustering issues. Since each task has two digits, the true number of clusters seen after training each task sequentially is 2, 4, 6, 8, 10. The number of clusters automatically detected by \acrshort{dbull} after training $\textrm{TASK}_1, \ldots$, $\textrm{TASK}_5$ is 4, 6, 8, 10, 12. \acrshort{dbull} clusters digit 1 into three clusters of different handwritten patterns in $\textrm{TASK}_1$. For other digits, \acrshort{dbull} discovers each new digit into one exact cluster as the ground truth. In contrast, CURL-D  clustered digits 0-1 into 14-16 clusters of $\textrm{TASK}_1$ and obtained 23-25 clusters for 10 digits after training five tasks sequentially. We provide visualization of the reconstructed cluster mean from the DP mixture model via our trained decoder of \acrshort{dbull} in Fig.~\ref{fig:cluster_mean}. 

Besides the overall clustering quality reported, we also provide the precision and recall of \acrshort{dbull} to view the performance for each digit after sequentially training all tasks. CURL-D overclusters the 10 digits into 25 clusters, making it hard to report the precision and recall of each digit. To visualize the results, the three sub-clusters of digit one by \acrshort{dbull} have been merged into one cluster. Overall, there is no significant performance loss of previous tasks after sequentially training multiple tasks for digits 0, 1, 3, 4, 6, 8, 9. Digit 2 has experienced precision decrease after training $\text{TASK}_4$ of digits 6 and 7 since \acrshort{dbull} has trouble in differentiating some samples from digits 2 and 7.


\begin{figure}[H]
\centerline{\includegraphics[scale=0.3]{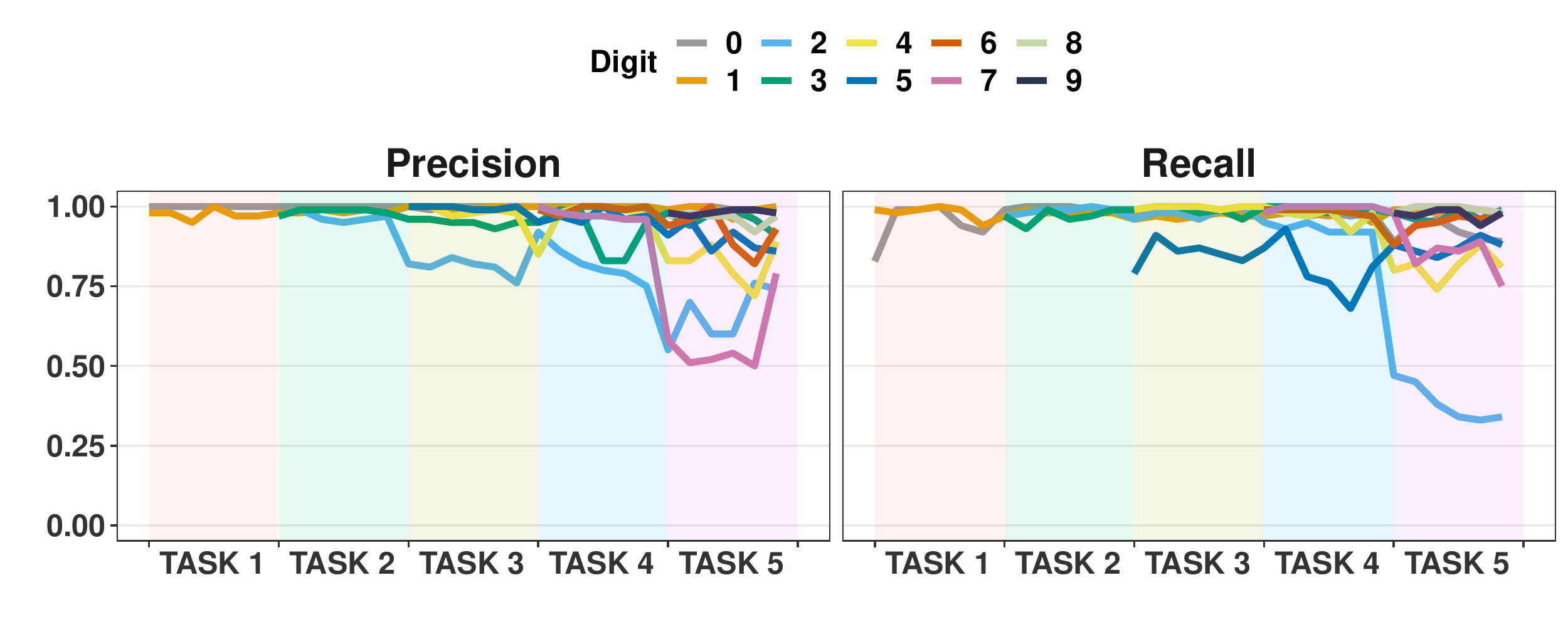}}
\caption{Precision and recall for each digit of DBULL evaluated after sequentially training from $\textrm{TASK}_1$ to $\textrm{TASK}_5$ across every 500 iterations for each task.}
\label{fig:prec_recall}
\end{figure}

\subsection{Batch Setting Clustering Performance Comparison}
\textbf{Experiment Objective.} The goal of this experiment is to demonstrate the generality our \acrshort{ll} method \acrshort{dbull}, which can achieve comparable performance as competing methods in an unsupervised batch setting. 

\textbf{Experiment Setup.} To examine our method performance in a batch setting, we test it on more complex datasets including \emph{Reuters10k} obtained from the original \emph{Reuters} \citep{lewis2004rcv1} and image \emph{STL-10} \citep{coates2011analysis}. We use the same Reuters10k and STL-10 dataset from \cite{xie2016unsupervised, jiang2017variational}. The details of Reuters10k and STL-10 are provided in Appendix B. For all datasets, we randomly select 80\% of the samples as training and evaluate the performance on the rest 20\% of the samples for all methods. 

\textbf{Discussion on Performance.} The true number of clusters is provided to competing methods: DEC, VaDE and CURL-F in advance since the total number of clusters is required. DBULL, CURL-D, VAE+DP have less information than DEC, VaDE and CURL-F since 
they have no knowledge about the true number of clusters. DBULL, CURL-D and VAE+DP all start with one cluster and detect the number of clusters on the fly. Thus, if DBULL can achieve similar performance to DEC, VaDE and CURL-F and outperforms its \acrshort{ll} counterpart CURL-D, it demonstrates \acrshort{dbull}’s effectiveness. Table~\ref{tb:multiple} shows \acrshort{dbull} performs the best in NMI, VM for MNIST and NMI, ARI and VM for STL10 and outperforms CURL-D in MNIST and STL10. Moreover, \acrshort{dbull} and DEC are more stable in terms of all evaluation metrics because of smaller standard error than other methods. We also report the number of clusters found by DBULL, CURL-D for MNIST, Reuters10k and STL-10 in Table~\ref{tb:clusters} out of five replications. Table~\ref{tb:clusters} shows that DBULL handles overclustering issues better in comparison with CURL-D. In summary, Table~\ref{tb:multiple} and  \ref{tb:clusters} demonstrate DBULL's effectiveness in a batch setting. If we fix the number of clusters as the true one for methods VaDE, DEC and CURL-F while training, the clustering accuracy can be considered as the classification accuracy. However, we are not able to set the number of clusters as the true one in DBULL. As we have seen in Table~\ref{tb:clusters}, out of five replications, the number of clusters found by DBULL is from 11 to 15. To compute a clustering accuracy for DBULL, taking MNIST as an example,  we count the number of correctly clustered samples from the biggest 10 clusters and divide it by the total number of samples in the testing phase. We report the accuracy comparison in Table~\ref{tb:cluster_acc}.


\begin{table}[p]
\caption{Clustering quality (\%) comparison averaged over five replications with both the average value and the standard error (in the parenthesis) provided.}
\label{tb:multiple}
\begin{center}
\scriptsize
\begin{tabular}{c | c c c }
    \toprule
    Dataset &  Method & NMI & ARI \\ [0.5ex]  
 \hline
\multirow{3}{*}{MNIST}& DEC & 84.67 (2.25) & \textbf{83.67} (4.53) \\
                        &VaDE &   80.35 (4.68) & 74.06 (9.11)  \\
                       & VAE+DP & 81.70 (\textbf{0.825}) & 70.49 (1.654)  \\
                        
        &CURL-F & 69.76 (2.51) & 56.47 (4.11)\\
        &CURL-D & 63.51 (1.32) & 36.84 (\textbf{1.98})\\
        &\textbf{DBULL} & \textbf{85.72} (1.02)& 83.53 (2.35)\\
        
\hline
\multirow{3}{*}{Reuters10k}& DEC & 46.56 (5.36) & 46.86 (7.98) \\ 
& VaDE & 41.64 (4.73) & 38.49 (5.44) \\
& VAE + DP & 41.62 (2.99) & 37.93 (4.57) \\
&CURL-F & \textbf{51.92} (3.22) & \textbf{47.72} (4.00)\\
&CURL-D & 46.31 (1.83) & 22.00 (\textbf{3.60})\\
& \textbf{DBULL} & 45.32 (\textbf{1.79}) & 42.66 (5.73)  \\
\hline
\multirow{3}{*}{STL10} & DEC & 71.92 (2.66) & 58.73 (5.09) \\
& VaDE & 68.35 (3.85) & 59.42 (6.84) \\
&VAE+DP & 43.18 (1.41) & 26.58 (1.32)  \\
&CURL-F & 66.98 (3.38) & 51.24 (4.06)\\
&CURL-D & 65.71 (1.33) & 37.96 (4.69)\\
&\textbf{DBULL} &\textbf{75.26} (\textbf{0.53}) & \textbf{70.72} (\textbf{0.81}) \\
\midrule

    Dataset &  Method & HS & VM \\ [0.5ex]  
 \hline
\multirow{3}{*}{MNIST}& DEC  & 84.67 (2.25) & 84.67 (2.25)   \\
                        &VaDE & 79.86 (4.93) & 80.36 (4.69)  \\
                       & VAE+DP  & \textbf{91.27} (\textbf{0.215}) & 81.19 (0.904)  \\
&CURL-F & 68.60 (2.56) & 69.75 (2.51)\\
&CURL-D & 76.35 (1.53) & 62.45 (1.32) \\
&\textbf{DBULL} & 89.34 (0.25)  & \textbf{85.65} (\textbf{0.51}) \\
\hline
\multirow{3}{*}{Reuters10k}& DEC & 48.44 (5.44) & 46.52(5.36)  \\ 
& VaDE &  43.64 (4.88) & 41.60 (4.73)\\
& VAE + DP & 46.64 (3.85) & 41.34 (2.94) \\
&CURL-D & \textbf{66.90} (2.09) & 43.34 (\textbf{2.00})\\
&CURL-F & 54.38 (3.49) & \textbf{51.86} (3.21)\\
& \textbf{DBULL} & 48.88 (\textbf{1.86})  & 45.40 (2.04)\\
\hline
\multirow{3}{*}{STL10} & DEC  & 68.47 (3.48) & 71.83 (2.72) \\
& VaDE & 67.24 (4.23) & 68.37 (3.92) \\
&VAE+DP  & 42.28 (\textbf{1.03}) & 43.16 (1.39) \\
&CURL-F & 65.46 (3.27) & 66.96 (3.37)\\
&CURL-D & \textbf{80.86} (2.94) & 64.31 (1.24)\\
&\textbf{DBULL} & 77.61 (1.29)  & \textbf{75.22} (\textbf{0.52}) \\
\bottomrule
\end{tabular}
\end{center}
\end{table}

\begin{table}[ht]
\scriptsize
\centering
\caption{Number of clusters found by DBULL and CURL-D out of five replications, where the upperbounds for the number of clusters for Reuters10k and STL-10 are set at 40 and 50.}
\begin{tabular}{cccc}
\toprule
    Datasets& 
        True \# of Clusters &
        DBULL &
        CURL-D  \\
\midrule
    MNIST & 
       10 &
        11-15 &
        34 
        \\
  Reuters10k &
        4 & 		
        5-10 & 
        40  
        \\ 
  STL-10 &
        10 &
        12-15 &
        50 
        \\
 
\bottomrule
\end{tabular}

\label{tb:clusters}
\end{table}

\begin{table}[ht]
	\scriptsize
	\centering
	\caption{Clustering accuracy for VaDE, DEC, CURL-F and DBULL. We report the best accuracy for DBULL since DEC \citep{xie2016unsupervised} and VaDE \citep{jiang2017variational} only report their best accuracy. CURL-F \citep{rao2019continual} reports both the best accuracy and the average accuracy with the standard error.}
	\begin{tabular}{cccccc}
			\toprule
			Datasets& DEC & VaDE &CURL-F (best)& CURL-F (average)& DBULL \\
			\midrule
			MNIST & 84.30\% & 94.46\% & 84\%& 79.38\% (4.26\%) & 92.27\%
			\\
			\bottomrule
	\end{tabular}
		
	\label{tb:cluster_acc}
\end{table}

\section{Conclusion}
In this work, we introduce our approach DBULL for unsupervised LL problems. DBULL is a novel   end-to-end   approximate Bayesian  inference  algorithm, which is able to perform automatic new task discovery  via  our proposed  dynamic  model  expansion  strategy, adapt  to  changes  in  the  evolving data distributions, and overcome forgetting using our proposed information extraction mechanism via summary sufficient statistics while learning the underlying representation simultaneously. Experiments on MNIST, Reuters10k and STL-10 demonstrate that DBULL has competitive performance compared with state-of-the-art methods in both a batch setting and an unsupervised LL setting.

\section*{Acknowledgment}
The work described was supported in part by Award Numbers  U01 HL089856 from the National Heart, Lung, and Blood Institute and NIH/NCI R01 CA199673.

\newpage
\bibliography{ref}
\newpage
\appendix
\section{Derivation of $\mathcal{L}_{\textrm{ELBO-VAE}}$}\label{app:derivation}
Here we provide the derivation of $\mathcal{L}_{\textrm{ELBO-VAE}}$ described in Section~\ref{sec:label}. The notations are defined in Table~\ref{tab:not}. 

Recall that in Section~\ref{sec:label}, the ELBO is 
\begin{align} 
            \begin{split}
&\mathbb{E}_{q(\bm{y}, \bm{z}, \bm{\phi}, \bm{v}|\bm{x})}
    \left[\log\frac{p_\theta(\bm{x}|\bm{z})p(\bm{z}|\bm{y}, \bm{\phi})p(\bm{y}|\bm{v})p(\bm{v})p(\bm{\phi})}{q_{\psi}(\bm{z}|\bm{x})q(\bm{y})q(\bm{v})q(\bm{\phi}))}\right]
    \\
    &= \mathbb{E}_{q(\bm{y}, \bm{z}, \bm{\phi}, \bm{v}|\bm{x})}
    \left[\log p_\theta(\bm{x}|\bm{z})\right]
    +\mathbb{E}_{q(\bm{y}, \bm{z}, \bm{\phi}, \bm{v}|\bm{x})}
    \left[\log p(\bm{z}|\bm{y}, \bm{\phi})\right]\\
    &- \mathbb{E}_{q(\bm{y}, \bm{z}, \bm{\phi}, \bm{v}|\bm{x})}
    \left[\log q_{\psi}(\bm{z}|\bm{x})\right] 
    \\
   &+ \mathbb{E}_{q(\bm{y}, \bm{z}, \bm{\phi}, \bm{v}|\bm{x})}
    \left[\log p(\bm{y}|\bm{v})\right] 
    + \mathbb{E}_{q(\bm{y}, \bm{z}, \bm{\phi}, \bm{v}|\bm{x})}
    \left[\log p(\bm{v})\right] 
    \\
    &- \mathbb{E}_{q(\bm{y}, \bm{z}, \bm{\phi}, \bm{v}|\bm{x})}
    \left[\log q(\bm{y})\right] 
    - \mathbb{E}_{q(\bm{y}, \bm{z}, \bm{\phi}, \bm{v}|\bm{x})}
    \left[\log q(\bm{v})\right] 
    \\
    &- \mathbb{E}_{q(\bm{y}, \bm{z}, \bm{\phi}, \bm{v}|\bm{x})}
    \left[\log q(\bm{\phi})\right] 
    + \mathbb{E}_{q(\bm{y}, \bm{z}, \bm{\phi}, \bm{v}|\bm{x})}
    \left[\log p(\bm{\phi})\right].
        \end{split}
\label{eq:a_all_elbo}
\end{align}
 $\mathcal{L}_{\textrm{ELBO-VAE}}$ only involves the first three terms of equation~\eqref{eq:a_all_elbo} that make a contribution to optimize $\theta$, $\psi$ and $\bm{z}$ since we adopt the alternating optimization technique. 
 
 We assume our variational distribution takes the form of 
\begin{align} 
     \begin{split}
&q(\bm{y}, \bm{z}, \bm{\phi}, \bm{v}|\bm{x})=q_{\psi}(\bm{z}|\bm{x})q(\bm{\phi})q(\bm{v})q(\bm{y})   \\
   &=\mathcal{N}(\bm{\mu}(\bm{x};\psi), \textrm{diag}(\bm{\sigma}^2(\bm{x}; \psi)))\prod_{t=1}^{T-1} q_{\eta_t}(v_t)\prod_{t=1}^T q_{\zeta_t}(\bm{\phi}_t)\prod_{n=1}^{N}q_{\rho_n}(\bm{y}_n)\\
    &=\mathcal{N}(\bm{\mu}(\bm{x};\psi), \textrm{diag}(\bm{\sigma}^2(\bm{x}; \psi))) \prod_{t=1}^{T-1}\textrm{Beta}(\eta_{t_1}, \eta_{t_2})\\
    &\prod_{t=1}^T\mathcal{N}(\bm{\mu}_t|\bm{m}_t, (\bm{\beta}_t\bm{\Lambda}_t)^{-1})\mathcal{W}(\bm{\Lambda}_t|W_t, \nu_t)\prod_{n=1}^{N} \textrm{Mult}(T, \rho_n),
    \end{split}
\label{eq:a_variational_new}
\end{align}
where we denote $f_\psi(\bm{x})=(\bm{\mu}(\bm{x};\psi), \bm{\sigma}^2(\bm{x}; \psi))$, which is a neural network, $T$ is the number of mixture components in the DP of the variational distribution, $\bm{z}_n\sim \mathcal{N}(\bm{z}_n|\bm{\mu}_t, \bm{\Lambda}_t^{-1})$, $\bm{\phi}_t=(\bm{\mu}_t, \bm{\Lambda}_t)$, and $\textrm{Mult}(T, \rho_n)$ is a Multinomial distribution. Under the assumptions in \eqref{eq:a_variational_new},  we derive each of the first three terms in Equation~\ref{eq:a_all_elbo} to obtain $\mathcal{L}_{\textrm{ELBO-VAE}}$.

\begin{table}[!htb]
  \caption{Notations in the ELBO.} 
 \label{tab:not}
   \small 
   \centering 
   \begin{tabular}{l} 
   \toprule[\heavyrulewidth]\toprule[\heavyrulewidth]
   \textbf{Notations in the ELBO} \\ 
   \midrule
      $N$: the total number of observations.\\
   $L$: the number of Monte Carlo samples  in Stochastic\\ Gradient
   Variational Bayes (SGVB).\\ 
   $\Sigma: \textrm{diag}(\bm{\sigma}^2(\bm{x};\psi))$. \\
   $\bm{x}_n$: the $n$th observation. \\
   $\bm{y}_n$: cluster membership for the $n$th observation.\\ 
   $p(\bm{y}_n=k) = \gamma_{ik}, N_k = \sum_{n=1}^N \gamma_{nk}$.  \\
   $\hat{\bm{z}}_n=\frac{1}{L}\sum_{l=1}^L \bm{z}_n^{(l)}.$ \\
   $\bar{\bm{z}}_k=\frac{1}{N_k}\sum_{n=1}^{N}\gamma_{nk}\hat{\bm{z}}_n$.\\
     $\bm{S}_k = \frac{1}{N_k}\sum_{n=1}^N \gamma_{nk} (\hat{\bm{z}}_n - \bar{\bm{z}}_k)(\hat{\bm{z}}_n - \bar{\bm{z}}_k)^T.$ \\
     $\beta_k = \beta_0+ N_k$: the scalar precision in NW distribution. \\
    $\bm{m}_k=\frac{1}{\beta_k}(\beta_0\bm{m}_0+N_k\bar{z}_k)$: the posterior mean of cluster $k$. \\
    $W_k^{-1} = W_0^{-1} + N_k S_k + \frac{\beta_0 N_k}{\beta_0 + N_k}(\bar{\bm{z}}_k- \bm{m}_0)(\bar{\bm{z}}_k- \bm{m}_0)^T$.\\ 
    $\nu_k = \nu_0+N_k$: the $k$th posterior degrees of freedom of NW. \\
    $\bm{\phi}$: variational parameters of the $k$th NW components.\\
   $\eta_t$: variational parameters of a Beta distribution for the\\
   $t$th component in Equation~\ref{eq:a_variational_new}.\\
  $\zeta_t$: variational parameters of the NW distribution for $\bm{\phi}_t$. \\
  $\rho_n$: the variational parameters of a categorical distribution \\
  for the cluster membership for each observation.\\
   \bottomrule[\heavyrulewidth] 
   \end{tabular}
\end{table}

(1) $\mathbb{E}_{q_{\psi}(\bm{z}|\bm{x})q(\bm{y})q(\bm{v})q(\bm{\phi})}\left[\log P_\theta(\bm{x}|\bm{z})\right]$:\\
We assume in the generative model that  $\bm{x}|\bm{z}=\bm{z} \sim \mathcal{N}\left(\bm{\mu}(\bm{z}; \theta), \textrm{diag}(\bm{\sigma}^2(\bm{z};\theta))\right)$ and $p_\theta(\bm{x}|\bm{z})$ is parameterized by a neural network $g_\theta: Z\rightarrow X$ and $g_\theta(\bm{z})=\left(\bm{\mu}(\bm{z}; \theta), \bm{\sigma}^2(\bm{z}; \theta)\right)$. Using the reparameterization trick \citep{kingma2014auto} and  the Monte Carlo estimate of expectations, we have   
\begin{align} 
  \mathbb{E}&_{q_{\psi}(\bm{z}|\bm{x})q(\bm{y})q(\bm{v})q(\bm{\phi})}[\log P_\theta(\bm{x}|\bm{z})]\notag\\
   & =\mathbb{E}_{q_{\psi}(\bm{z}|\bm{x})q(\bm{y})q(\bm{v})q(\bm{\phi})}\bigg(-\frac{1}{2}\log(\bm{\sigma}^2(\bm{z}; \theta)_j)
    +\frac{(\bm{x}_j-\bm{\mu}(\bm{z}; \theta)_j)^2}{\bm{\sigma}^2(\bm{z}; \theta)_j}    \bigg)
    \notag\\
    &=-\frac{1}{2}\frac{1}{L}\sum_{l=1}^L\sum_{i=1}^N\sum_{j=1}^D\bigg(\log(\bm{\sigma}^2(\bm{z}; \theta)_j^{(l)}+ 
    \frac{\left(\bm{x}_{ij}- \bm{\mu}(\bm{z}; \theta)_j^{(l)}\right)^2}{\bm{\sigma}^2(\bm{z}; \theta)_j^{(l)}}\bigg).
        \label{eq:3}
\end{align}

(2) $\mathbb{E}_{q_{\psi}(\bm{z}|\bm{x})q(\bm{y})q(\bm{v})q(\bm{\phi})}[\log p(\bm{z}|\bm{y}, \bm{\phi})]:$ 

Recall that $\psi$,
where $q_\psi(\bm{z}|\bm{x})= \mathcal{N}(\bm{\mu}(\bm{x};\psi), \bm{\sigma}^2(\bm{x}; \psi))$ and $
        (\bm{\mu}(\bm{x};\psi), \bm{\sigma}^2(\bm{x}; \psi))=f(\bm{x}; \psi),$ where $f$ is a neural network.  Following \cite{kingma2014auto}, we use the reparameterization and sampling trick to allow backpropagation, for $l=1,2, \ldots, L,$ where $L$ is the number of Monte Carlo samples, we have
\begin{equation*}
\bm\epsilon^{(l)}\sim\mathcal{N}(\bm{0}, \bm{I}) ~\ \textrm{and} ~\ \bm{z}^{(l)}= \bm{\mu}(\bm{x};\psi)+\bm{\epsilon}^{(l)}\bm{\sigma}(\bm{x}; \psi).
\end{equation*}
Define $
    \hat{\bm{z}}_n =  \frac{1}{L}\sum_{l=1}^L \bm{z}_n^{(l)}.
$
We have
\begin{align} 
\begin{split}
&\mathbb{E}_{q_{\psi}(\bm{z}|\bm{x})q(\bm{y})q(\bm{v})q(\bm{\phi})}[\log p(\bm{z}|\bm{y}, \bm{\phi})]\notag
\\ &=\frac{1}{2}\sum_{k=1}^T N_k\left\{
\log \tilde{\Lambda}_k - D\beta_k^{-1} - \nu_k\mbox{Tr}(S_k W_k)\right\}\notag\notag\\
&-\frac{1}{2}\sum_{k=1}^T N_k\left\{\nu_k(\bar{z}_k -\bm{m}_k)^T W_k(\bar{z}_k -\bm{m}_k)- D\log(2\pi)
\right\},
\end{split}
\end{align} 
where
\begin{align} 
  \label{third}
  \begin{split}
   \log\tilde{\Lambda}_k = \mathbb{E}[\log\Lambda_k]
   = \sum_{j=1}^D\psi\left(\frac{\nu_k+1-i}{2}\right) + D\log 2 + \log\vert W_k\vert.
  \end{split}
\end{align}

(3) $\mathbb{E}{q_{\psi}(\bm{z}|\bm{x})q(\bm{y})q(\bm{v})q(\bm{\phi})}(\log q_{\psi}(\bm{z}|\bm{x}))$:\\
Since $q_\psi(\bm{z}|\bm{x})=  \mathcal{N}(\bm{\mu}(\bm{x};\psi), \textrm{diag}(\bm{\sigma}^2(\bm{x}; \psi)))$, $\mathbb{E}{q_{\psi}(\bm{z}|\bm{x})q(\bm{y})q(\bm{v})q(\bm{\phi})}(\log q_{\psi}(\bm{z}|\bm{x}))$ is equal to the negative entropy of a multivariate Gaussian distribution:
\begin{align} 
   \mathbb{E}{q_{\psi}(\bm{z}|\bm{x})q(\bm{y})q(\bm{v})q(\bm{\phi})}(\log q_{\psi}(\bm{z}|\bm{x})) 
= \frac{1}{2}\log(\textrm{Det}(2\pi e \Sigma)),
\end{align}
where $\Sigma = \textrm{diag}(\bm{\sigma}^2(\bm{x};\psi))$.

When we update the neural network parameters $\theta$ and $\psi$ and the latent representation $\bm{z}$, the DPMM parameters will be fixed. Thus,
the terms that do not involve $\bm{z}$, $\theta$, $\psi$ will not contribute to the $L_{\textrm{ELBO}-\textrm{VAE}}$. Hence, 
\begin{align}
&\mathcal{L}_{\textrm{ELBO}-\textrm{VAE}}(\bm{x}) 
= -\frac{1}{2} \sum_{k=1}^{T} N_k \nu_k \left\{ \textrm{Tr}(S_k W_k)\right\}\notag\\&-\frac{1}{2} \sum_{k=1}^{T} N_k \nu_k\left\{(\bar{z}_k-\bm{m}_k)^T W_k (\bar{z}_k - \bm{m}_k)\right\} \notag\\
&-\frac{1}{2}\frac{1}{L}\sum_{l=1}^L\sum_{i=1}^N\sum_{j=1}^D\bigg(\log(\bm{\sigma}(\bm{z}, \theta)^2)_j^{(l)}+ \frac{
    \left(\bm{x}_{ij}- \bm{\mu}(\bm{z};\theta)_j^{(l)}\right)^2}{(\bm{\sigma}(\bm{z};\theta)^2)_j^{(l)}}\bigg)\notag\\&+\frac{1}{2}\log(\textrm{Det}(2\pi e \Sigma)).
\label{eq:a_elbo_derivation}
\end{align}
  
Details of the standard variational inference for parameters in the \acrshort{dp} Gaussian mixture can be found in \cite{blei2006variational}. We only list the updating equations for the variational parameters below. 

\begin{itemize}
    \item $q(\bm{y}_n=i)=\gamma_{n, i}.$ 
    \item $q(\bm{y}_n>i)= \sum_{j=i+1}^{T}\gamma_{n, j}.$
    \item $\mathbb{E}_{q}[\log V_i]=\Psi(\gamma_{i, 1})-\Psi(\gamma_{i, 1} + \gamma_{i, 2}).$
    \item $\mathbb{E}_{q}[\log(1-V_i)]=\Psi(\gamma_{i, 2}) - \Psi(\gamma_{i, 1} + \gamma_{i, 2}).$
        \item $S_t = \mathbb{E}[\log V_i] + \sum_{i=1}^{t-1}\mathbb{E}_q[\log(1-V_i)]+
    \frac{1}{2}\log\tilde{\Lambda}_k-\frac{D}{2\beta_k}-\frac{\nu_k}{2}(\hat{z}_n-\bm{m}_k)^T W_k (\hat{z}_n-\bm{m}_k).$
    \item $\gamma_{n, t}\propto \exp(S_t)$, $\gamma_{n, t} = \frac{\exp(S_t)}{\sum_{t=1}^T \exp(S_t)}$.
    \item Under the Normal-Wishart variational distribution assumption, \begin{align*} \label{six}
\begin{split}
&\mathbb{E}{q_{\psi}(\bm{z}|\bm{x})q(\bm{y})q(\bm{v})q(\bm{\phi})}(\log p(\bm{\phi})) 
    \\
&=
    \frac{1}{2}\sum_{k=1}^T \left\{
        D\log(\beta_0/2\pi) + \log \tilde{\Lambda}_k\right\}\notag\\
&-\frac{1}{2}\sum_{k=1}^T\left\{\frac{D\beta_0}{\beta_k} +\beta_0\nu_k(\bm{m}_k-\bm{m}_0)^T W_k (\bm{m}_k-\bm{m}_0)\right\} \\
        &+ T \log\mbox{B}(W_0, \nu_0)  \frac{\nu_0-D-1}{2}\sum_{k=1}^T \log\tilde{\Lambda}_k-\frac{1}{2}\sum_{i=1}^T \nu_k\mbox{Tr}(W_0^{-1} W_k),
    \end{split}
\end{align*}
        where 
\begin{align*}
&\mbox{B}(W, \nu) = \vert W\vert ^{-\nu/2} \left(2 ^{\nu D/2}\pi^{D(D-1)/4}\prod_{i=1}^{D}\Gamma\left(\frac{\nu+1-i}{2}\right)   \right)^{-1}.
\end{align*}
  \item Similarly, we have
  \begin{align*}
            \begin{split}
  & \mathbb{E}{q_{\psi}(\bm{z}|\bm{x})q(\bm{y})q(\bm{v})q(\bm{\phi})}[\log q(\bm{\phi})]\\
   &=
    \sum_{k=1}^T \left(  
        \frac{1}{2}\log \tilde{\Lambda}_k +\frac{D}{2}\log\left(\frac{\beta_k}{2\pi}\right)-\frac{D}{2}- H[q(\Lambda_k)]
    \right),
   \end{split}
\end{align*}

\begin{equation*}
    H[\Lambda] = -\log B(W, \nu)-\frac{\nu-D-1}{2}
    \mathbb{E}[\log |\Lambda|]+ 
   \frac{\nu D}{2}.
\end{equation*}
\end{itemize}
\newpage
\section{Dataset Description}

\textbf{Reuters10k.} We are using the same Reuters10k dataset from \cite{xie2016unsupervised, jiang2017variational}. The original Reuters dataset has 810000 English news stories labeled with a category tree. Following DEC and VaDE, we use 4 root categories: corporate/industrial, government/social, markets, and economics as labels and discard all documents with multiple labels and obtain 685071 articles. We computed tf-idf features on the top 2000 most frequently occurring words to represent all articles. Since some algorithms do not scale to the full Reuters dataset, a randomly 10000 examples are sampled and is referred to as Reuters10 in \cite{xie2016unsupervised}.

\textbf{STL-10.} The original STL-10 dataset consists of color images of dimension 96-by-96. There are 10 classes with a total of 1300 samples. Following \cite{jiang2017variational}, we extract features of STL-10 images using ResNet-50 \citep{he2016deep} to train and test the performance of DBULL and all other competing methods. To be specific, we applied a $3\times 3$ average polling over the last feature map of ResNet-50. 

\newpage
\section{Tuning Parameters and Implementation Details}\label{app:implementation}
\subsection{Hyperparameters in the \acrshort{dpmm}s}
In this paper, we choose the values of the hyperparameters in the DP following the common choices in the nonparametric Bayesian literature.

As we have discussed in Section~\ref{sec:dp}, in a \acrshort{dp}, the mixture proportions are drawn from a stick-breaking process as
\[
\pi_k=\begin{cases}
v_1, & \textrm{if}\quad k=1, \\
v_k\prod_{j=1}^{k-1}(1-v_j), & \textrm{for}\quad k>1,
\end{cases}
\]
where $v_k\sim \textrm{Beta}(1, \alpha)$. Here, $\alpha$ is a hyperparameter and we set $\alpha=1.0$. Under our Dirichlet Process Gaussian mixture model assumption, we choose the base distribution $G_0$ to be a Normal-Wishart distribution denoted as $G_0=\textrm{NW}(\boldsymbol{\lambda}_0)$ to generate the parameters for each Gaussian mixture, where $\bm{\lambda}_0=(m_0, \beta_0, \nu_0, W_0)=(\bm{0}, 0.2, D+2, I_{D\times D})$, and $D$ is the dimension of the latent vector $\bm{z}$. The values of the hyper-parameter $\bm{\lambda}_0$ are conventional choices in the  literature for Gaussian mixture. 

In Section~\ref{sec:expansion}, while performing the cluster expansion and redundancy removal strategy, we cache the samples in the subsample if the probability of the $n$th observation to be assigned to cluster $k$ is bigger than a threshold of value 0.1. We choose the value of 0.1 since \cite{hughes2013memoized} have suggested this value while performing the birth and merge moves for clustering in the nonparametric Bayesian context.

\subsection{Tuning Parameters in DNNs and Implementation Details}
In order to compare fairly with state-of-the-art methods such as VaDE and DEC, we adopted the same neural network architecture as DEC and VaDE. The pipeline is $d-500-500-2000-l$ and $l-2000-500-500-d$ for the encoder and decoder, respectively, where $d$ and $l$ denote the dimensionality of the input and latent features. All layers are fully connected and a sampling layer bridges the encoder and decoder. We adopt the same pre-trained Stacked Autoencoder in VaDE as the initialization for the neural network. Adam optimizer \citep{kingma2014adam} is used as the optimization engine to update the neural network. The batch size is set to 1500 for a batch setting. The size of each data stream is set to 1000 and we divide it into two mini-batches. The learning rate for Reuters-10K and STL-10 is set as 0.002 and 0.0002 for MNIST with a common decay rate of 0.9 for every epoch in a batch setting. For the clustering initialization, DEC and VaDE start with K-means or GMM and fix the number of clusters as the ground truth. DBULL starts with one cluster and grows or merges clusters using the cluster expansion and redundancy removal technique. In a lifelong learning setting, the fixed number of samples for replay is 100.

\end{document}

%% file: head_zhu.tex
\usepackage{amsmath,amssymb,amsfonts,amsthm}
\usepackage{bm}
\usepackage{graphicx}

\usepackage{float}
\usepackage{algorithm}
\floatstyle{ruled}
\newfloat{algo}{tbp}{loa}
\providecommand{\algorithmname}{Algorithm}
\floatname{algo}{\protect\algorithmname}

\DeclareRobustCommand{\ov}[1][\qedsymbol]{%
	\ifmmode \mathqed
	\else
	\leavevmode\unskip\penalty9999 \hbox{}\nobreak\hfill
	\quad\hbox{$\blacksquare$}%
	\fi
}

